\documentclass{article}

\usepackage[preprint, nonatbib]{arxiv}

\usepackage[utf8]{inputenc} 
\usepackage[T1]{fontenc}    
\usepackage{hyperref}       
\usepackage{url}            
\usepackage{booktabs}       
\usepackage{amsfonts}       
\usepackage{nicefrac}       
\usepackage{microtype}      
\usepackage{xcolor}         

\usepackage{amsmath}
\usepackage{bbm}
\usepackage{multirow}
\usepackage{colortbl}

\usepackage{array}    
\usepackage{caption}  
\usepackage{float}
\usepackage{pifont}
\usepackage{makecell}
\newcommand{\cmark}{\ding{51}}%
\newcommand{\xmark}{\ding{55}}%
\usepackage[ruled,vlined]{algorithm2e}
\usepackage{arydshln}

\usepackage{subcaption}
\usepackage[capitalize]{cleveref}
\usepackage{amsthm}
\usepackage{xspace}
\usepackage{graphicx} 
\usepackage{wrapfig}
\newcommand{\name}{MotionWavelet\xspace}

\newcommand{\x}{\mathbf{x}}
\newcommand{\y}{\mathbf{y}}
\newcommand{\R}{\mathbb{R}}

\makeatletter
\DeclareRobustCommand\onedot{\futurelet\@let@token\@onedot}
\def\@onedot{\ifx\@let@token.\else.\null\fi\xspace}
\def\aka{\emph{a.k.a}\onedot}

\def\ie{\emph{i.e}\onedot}

\def\vs{\emph{vs}\onedot}

\title{\name: Human Motion Prediction via Wavelet Manifold Learning}

\author{Yuming Feng$^{1\dag}$ \quad Zhiyang Dou$^{2,3,\dag,\ddag}$ \quad  Ling-Hao Chen$^{4}$ \quad Yuan Liu$^{5,6}$ \quad Tianyu Li$^7$
\And
Jingbo Wang$^{8}$ \quad Zeyu Cao$^{9}$ \quad Wenping Wang$^{10}$ \quad  Taku Komura$^2$ \quad Lingjie Liu$^{3,\ddag}$
}

\crefname{theorem}{Theorem}{Theorem}
\crefname{lemma}{Lemma}{Lemma}
\crefname{remark}{Remark}{Remark}
\crefname{figure}{Fig.}{Fig.}
\crefname{section}{Sec.}{Sec.}
\crefname{equation}{Eq.}{Eq.}
\crefname{table}{Tab.}{Tab.}
\crefname{algorithm}{Alg.}{Alg.}

\newtheorem{definition}{Definition}

\newcommand{\myPara}[1]{\vspace{.05in}\noindent\textbf{#1}}

\begin{document}

\maketitle

\let\thefootnote\relax\footnotetext{
$^1$Imperial College London, $^2$The University of Hong Kong, $^3$University of Pennsylvania  $^4$Tsinghua University, $^5$Hong Kong University of Science and Technology, $^6$Nanyang Technological University, $^7$Georgia Institute of Technology, $^8$Shanghai AI Lab, $^9$University of Cambridge, $^{10}$Texas A\&M University.

\quad Project Page: \url{https://frank-zy-dou.github.io/projects/MotionWavelet/index.html}

\quad $\dag, \ddag$ denote equal contributions and corresponding authors.}

\begin{abstract}

Modeling temporal characteristics and the non-stationary dynamics of body movement plays a significant role in predicting human future motions. However, it is challenging to capture these features due to the subtle transitions involved in the complex human motions. This paper introduces \name, a human motion prediction framework that utilizes Wavelet Transformation and studies the human motion patterns in the spatial-frequency domain. In \name, a \textit{Wavelet Diffusion Model}~(WDM) learns a \textit{Wavelet Manifold} by applying Wavelet Transformation on the motion data therefore encoding the intricate spatial and temporal motion patterns. Once the \textit{Wavelet Manifold} is built, WDM trains a diffusion model to generate human motions from Wavelet latent vectors. In addition to the WDM, \name also presents a \textit{Wavelet Space Shaping Guidance} mechanism to refine the denoising process to improve conformity with the manifold structure. WDM also develops \textit{Temporal Attention-Based Guidance} to enhance the prediction accuracy. Extensive experiments validate the effectiveness of \name, demonstrating improved prediction accuracy and enhanced generalization across various benchmarks. Furthermore, we showcase the capability of our method in different controllable motion prediction tasks. Our code and models will be released upon acceptance.
\end{abstract}

\section{Introduction}
\label{sec:intro}

Human Motion Prediction~(HMP)~\cite{chen2023humanmac, wei2023human, mao2019learning, wan2022learn, yuan2019diverse, yuan2020dlow, zhang2021we, fragkiadaki2015recurrent, dang2022diverse, xu2022diverse, mao2021generating, bhattacharyya2018accurate, barquero2023belfusion, sun2023towards} is a fundamental problem in computer vision, graphics, and robotics given its wide-ranging applications, such as autonomous driving~\cite{paden2016survey, ridel2018literature,kim2020pedestrian}, virtual reality~\cite{hou2019head,gamage2021so,kim2020pedestrian}, and human-robot interaction~\cite{gaomulti,wan2022learn, cheng2019purposive,koppula2013anticipating, lafleche2018robot,xia2018gibson}. Accurately predicting future human motion based on observed data is challenging due to the high complexity, variability, and stochastic nature of human motion.

Previous attempts~\cite{yuan2019diverse, yuan2020dlow, wan2022learn,dang2022diverse} typically directly use the high dimensional motion data in capturing the complex temporal and spatial dynamics of human movement, which hinders their performance in motion prediction with the sparse observational data. 
Considering this challenge, there is considerable research on neuromechanics of human motions~\cite{dimitrijevic1998evidence, yuste2005cortex,harkema2011effect, enoka2008neuromechanics,schneider2024muscles} verify that human motion is intrinsically linked to the frequency domain. These key observations motivate some methods to model human motion~\cite{holden2017phase, starke2022deepphase, zhang2018mode,chen2023humanmac, wei2023human} in the frequency domain to synthesize or predict motions. This frequency decomposition fashion of motion simplifies the capture of periodic movement features, ultimately enhancing predictive accuracy in the modeling process of human motion. 

Despite the progress in motion frequency modeling, these methods still struggle to model details of the motion sequences. Specifically, previous phase-based representations in the frequency domain~\cite{starke2019neural, holden2017phase, starke2022deepphase, mao2019learning,starke2023motion, li2024walkthedog} take the Fourier phase components to model temporal dynamics. However, windowed Fourier analysis struggles to capture non-stationary signals due to its limited window size, impeding its adaptability to capture local frequency variations and handling abrupt, non-stationary signals. Besides, existing DCT-based methods~\cite{chen2023humanmac, sun2023towards, barquero2023belfusion, wei2023human} directly eliminate high-frequency components in the frequency domain, overlooking critical details and diminishing predictive accuracy.

To resolve these issues, we introduce \name, a novel data-driven method for human motion prediction that utilizes a \emph{Motion Wavelet Manifold} derived from human motion data and a \emph{Wavelet Diffusion Model} for motion prediction. Different from previous phase-based representations in the frequency domain, the motion wavelet manifold models the motion sequence as a whole, excelling in modeling dynamic transitions where periodicity assumptions often break down (see~\cref{sec:wavelet_vs_phase}). Besides, in contrast to the DCT-based frequency methods of ignoring high-frequency motion signals, the motion wavelet manifold models both high-frequency and low-frequency signals along the temporal and spatial axes, explicitly. As a result, the wavelet manifold offers enhanced adaptability to varying frequency changes, enabling the capture of both local temporal characteristics and non-stationary dynamics. This detailed modeling allows \name to effectively represent subtle transitions and intricate movements, providing a robust foundation for predicting complex human motion patterns. Additionally, we provide a comprehensive study on the influence of different wavelet bases for human motion representation, offering valuable guidance for motion embedding using the wavelet transform.

Once we define the learned wavelet manifold, our approach leverages this manifold to predict future motions based on limited short-term movement observations. We introduce a \emph{Wavelet Motion Diffusion} (WDM) model, wherein the diffusion model effectively captures the motion distribution in the wavelet domain, enabling accurate human motion prediction. Technically, during iterative denoising, previous methods~\cite{chen2023humanmac, sun2023towards, barquero2023belfusion, wei2023human} often apply sequential denoising directly to noisy inputs, which are misaligned with underlying manifold structures and reduce predictive accuracy. In contrast, in this work, we do not use the mask-completion fashion and predict motions with observed motion wavelets classifier-free guidance. To additionally improve the guidance controlablity, we propose \textit{Wavelet Manifold Shaping Guidance} to enhance the prediction precision. 
Specifically, \textit{Wavelet Manifold Shaping Guidance} map the output of denoiser via an Inverse DWT to recover motion signals, then reapply a DWT operation returning it into the wavelet manifold. Aligning the latent space with wavelet manifolds enables a structured progression of denoising steps, improving the quality and fidelity of generated results while stabilizing latent space noise and smoothing transitions across the diffusion trajectory, which is shown in~\cref{sec:wmsg_guidance}. Besides, we delve into the self-attention mechanism of the noise prediction network and find the self-attention mechanism mainly mines the motion coherence in the temporal dimension. Based on this observation, we propose \textit{Temporal Attention-Based Guidance} to further enhance predictive accuracy during the denoising process. By adaptively weighting the influence of different time steps according to the attention score, this approach emphasizes critical wavelet features while suppressing irrelevant noise (see Sec.~\ref{sec:tabg_guidance}), enabling a more accurate alignment with the target motion trajectory.

We conduct extensive experiments to demonstrate the effectiveness of our method. \name achieves high predictive accuracy and generalizability across a broad range of benchmarks, highlighting its capability to handle complex motions and diverse movement styles. A comprehensive evaluation and analysis are presented to validate the effectiveness of all key design components in \name.
\vspace{-0.6em}
\section{Related Work}
\noindent\textbf{Human Motion Prediction.}
Extensive studies~\cite{fragkiadaki2015recurrent, li2017auto, gurumurthy2017deligan, yuan2019diverse, salzmann2022motron, bhattacharyya2018accurate, yuan2020dlow, mao2021generating, zhang2021we, dang2022diverse, xu2022diverse, wei2023human, barquero2023belfusion, chen2023humanmac, sun2023towards} have been conducted in the field of Human Motion Prediction~(HMP). Various model architectures have been explored in these studies. For instance, some earlier works~\cite{fragkiadaki2015recurrent, li2017auto, bhattacharyya2018accurate} utilize Long Short-Term Memory (LSTM) networks~\cite{graves2013generating}, while MOJO~\cite{zhang2021we} applies the Gated Recurrent Unit (GRU)~\cite{cho2014learning}, and Transformer structures~\cite{vaswani2017attention} have also been explored in this context~\cite{liu2021multimodal, wei2023human}. Many of these approaches follow an AutoEncoder paradigm~\cite{yuan2019diverse, salzmann2022motron, blattmann2021behavior, xu2022diverse, yuan2020dlow, zhang2021we, dang2022diverse}, where the motion data is mapped into a latent space to facilitate the prediction task. Generative Adversarial Networks (GANs) have been utilized by framing the prediction task as a generative process~\cite{gurumurthy2017deligan}. More recently, diffusion-based models~\cite{sohl2015deep, song2020denoising, ho2020denoising} have been applied in human motion prediction~\cite{chen2023humanmac, sun2023towards, barquero2023belfusion, wei2023human}. In contrast to existing methods, we propose a novel approach that leverages wavelet manifold learning to enable wavelet manifold diffusion for motion prediction, utilizing the learned wavelet manifold.

\noindent\textbf{Motion in Frequency Domain.} Given that human body movement is intrinsically linked to the frequency domain, as evidenced by neuroscience findings, e.g., neuromechanics of human movement~\cite{dimitrijevic1998evidence, yuste2005cortex,harkema2011effect, enoka2008neuromechanics}, frequency domain approaches have been proposed for a variety of motion-related tasks~\cite{chen2023humanmac, mao2021generating, mao2019learning, unuma1995fourier, bruderlin1995motion, liu1994hierarchical, starke2020local, yumer2016spectral}, such as motion editing~\cite{bruderlin1995motion} and motion generation~\cite{liu1994hierarchical, starke2022deepphase, holden2017phase}. Inspired by the previous efforts, in this paper, we present the first application of wavelet manifold diffusion for motion prediction, enhancing guidance for wavelet manifold denoising.

\noindent\textbf{Denoising Diffusion Model.} Denoising Diffusion models~\cite{sohl2015deep, song2020denoising, ho2020denoising} have recently demonstrated impressive performance across various generation tasks, including 2D image synthesis~\cite{zhang2023adding, rombach2022high, saharia2022photorealistic}, 3D shape generation~\cite{lin2023magic3d, long2024wonder3d, yu2025surf, wang2023disentangled, zhang2024clay, tang2025lgm}, and motion synthesis~\cite{shafir2023human,luhumantomato, zhou2025emdm, chen2023executing, huang2025controllable, wan2023tlcontrol, zhang2022motiondiffuse,chen2024motionclr,dai2025motionlcm}. In human motion prediction, diffusion models have also been employed. For instance,~\cite{blattmann2021behavior} introduces a latent diffusion model tailored for behavior-driven human motion prediction, while MotionDiff~\cite{wei2023human} utilizes a spatial-temporal transformer-based diffusion network to generate diverse and realistic motions, with a graph convolutional network refining the outputs. More recently, HumanMAC~\cite{chen2023humanmac} adopts a diffusion model for motion prediction in a masked completion fashion. Unlike these approaches in HMP, we propose Wavelet Manifold Diffusion with novel designs to guide denoising and enhance alignment with the modeling of wavelet manifold.

\section{Method}

\subsection{Preliminaries}
\label{sec:preliminary}

\begin{figure}
    \centering
\includegraphics[width=\textwidth]{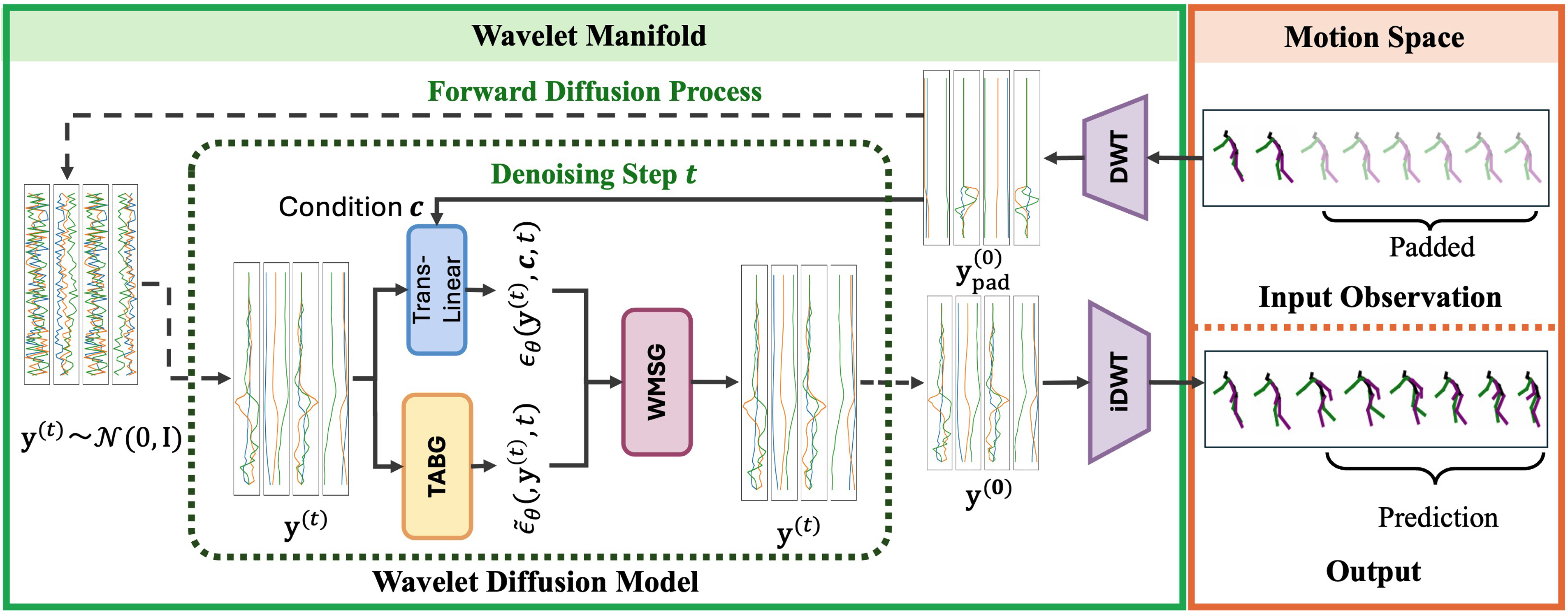}
    \caption{System overview. Our method first converts motion from spatial space to Wavelet manifold~(Sec.~\ref{sec:waveletmanifold}) and then conducts Wavelet Manifold Diffusion~(Sec.~\ref{sec:wmd}) given few history frames where a denoiser $\epsilon_\theta$ is trained from the diffusion process $q (\mathbf{y}^{(t)}| \mathbf{y}^{(t-1)})$. During inference, the Wavelet Manifold Diffusion model predicts the latent $\mathbf{y}^{(0)}$ from condition inputs and then uses $\mathtt{iDWT}$ to transform it to the motion space efficiently. }
    \label{fig:pipeline}
\end{figure}

\myPara{Problem Definition.} We denote the sequence of motion history of $H$ frames as $\mathbf{x}^{1:H}\in \R^{H \times 3J}$, where $\mathbf{x}^i \in \R^{3J}$ is defined as the pose at time-step $i$ with number of body joints $J$. The objective of the Human Motion Prediction (HMP) task consists in predicting the following future $F$ frames of motions $\x^{H+1:H+F}\in \R^{F \times 3J}$ given a motion history $\x^{1:H}$.

\myPara{Discrete Wavelet Transform Formulation.} The vanilla Discrete Wavelet Transform (\aka DWT) is designed to capture the high-frequency and low-frequency components along the whole sequence. The DWT process involves two steps: convolution filtering and down-sampling. 

The DWT decomposes a discrete sequence signal $x[n]$ with length $N$ into low-frequency coefficients and high-frequency coefficients. Given a mother wavelet $\psi[n]$ and its corresponding scaling function $\phi[n]$\footnote{Here, $\psi[n]$ and $\phi[n]$ satisfy $\psi_k[n]=2^{-1/2} \, \phi[2^{-1} n - k]$ and $\phi_k[n]=2^{-1/2} \, \psi[2^{-1} n - k]$, according to the Dilation Equations.}, $x[n]$ is represented as a linear combination of wavelet basis and scaling factors,
\begin{equation}
    x[n] = \sum\nolimits_{k} a[k] \, \phi_{k}[n] + \sum\nolimits_k d[k] \, \psi_{k}[n],
    \label{eq:1d-dwt}
\end{equation}
where $a[k]$s are low-frequency coefficients and $d[k]$s are high-frequency coefficients, respectively. Accordingly, the decomposed coefficients can be formulated as, 
\begin{equation}
    a[k] = \sum\nolimits_{n} x[n] \, \phi_{k}[n], \ \
    d[k] = \sum\nolimits_{n} x[n] \, \psi_{k}[n].
    \label{eq: coff}
\end{equation}
Specifically, the coefficients in~\cref{eq: coff} are practically obtained by discrete convolutions with down-sampling, 
\begin{equation}
\small
    a[k] = \sum\nolimits_{n}h[n]\cdot x[2k-n], \ \
    d[k] = \sum\nolimits_{n}g[n]\cdot x[2k-n],
\end{equation}
where coefficients $\psi[n]$ and $\phi[n]$ are further defined as $\phi[n] = \sqrt{2}\sum_k h[k] \, \phi[2n - k]$, and $\psi[n] = \sqrt{2}\sum_k g[k] \, \phi[2n - k]$. Here, $h[k]$ and $g[k]$ are the low-pass and high-pass filters, respectively. Consequently, the original signal $x[n]$ can be obtained with low-frequency $a[n]$ and high-frequency $d[n]$ coefficients by applying~\cref{eq:1d-dwt}.

\subsection{Wavelet Diffusion Model}
\subsubsection{Motion Wavelet Manifold}
\label{sec:waveletmanifold}

As discussed in~\cref{sec:preliminary}, a motion sequence can be represented as a 2-dimensional tensor, \ie $\mathbf{x}^{1:H+F}\in \R^{(H+F) \times 3J}$. Accordingly, we apply vanilla DWT operation to the human motion by applying the wavelet transform along both temporal and spatial dimensions. Technically, given a motion sequence $\mathbf{x}[i, j]$ ($i \in [1, H + F], j \in [1, 3J]$), the 2-D DWT decomposes it into four subbands $\mathbf{x}_{h,v}[k_1,k_2]$, where $h,v \in \{L,H\}$ denote low-pass or high-pass filter applied along horizontal or vertical directions, respectively. Technically, the coefficients are computed as,

\begin{equation}
\mathbf{y}_{h,v}[k_1,k_2] = \sum_{i}\sum_{j} f_h[i - 2k_1]\,f_v[j - 2k_2]\,\mathbf{x}[i, j],
\end{equation}
where $f_L[n] = h[n]$ and $f_H[n] = g[n]$ are the low-pass and high-pass filters from the 1-D DWT. For each subband of frequencies $\mathbf{y}_{h,v} \in \mathbb{R}^{K\times D}$, $K = \left\lfloor \frac{H + F + l - 1}{2} \right\rfloor$, $D = \left\lfloor \frac{3J + l - 1}{2} \right\rfloor$, where $l$ is the length of the filter. Thus $\mathbf{y}_{L, L}$ is the approximation coefficients, $\mathbf{y}_{L, H}$ and $\mathbf{y}_{H, L}$ are temporal and spatial detail coefficients respectively, and $\mathbf{y}_{H, H}$ corresponds to the spatial-temporal detail coefficients.

\begin{definition}[Motion Wavelet Manifold]
Given a motion $\mathbf{x}^{1:H+F}\in \R^{(H+F) \times 3J}$, we concatenate four subbands together $\mathbf{y}=\mathtt{cat}(\mathbf{y}_{L, L}, \mathbf{y}_{H, L}, \mathbf{y}_{L, H}, \mathbf{y}_{H, H}) = \mathtt{DWT}(\mathbf{x})\in \mathbb{R}^{K \times 4D}$ as the motion wavelet manifold. 
\end{definition}

This approach benefits motion generation by allowing the model to leverage both high- and low-frequency information explicitly, thereby improving the granularity of the predictions. The inverse DWT process is accordingly denoted as $\mathbf{x} = \mathtt{iDWT}(\mathbf y)$, which is similar to the 1-D DWT in~\cref{sec:preliminary}.

Based on the DWT transformation, we introduce the proposed motion wavelet manifold as follows. Given a complete motion sequence $\mathbf{x} \in \mathbb{R}^{(H+F) \times 3J}$ and a wavelet function $\psi$, we apply the DWT to decompose $\mathbf{x}$ into subbands $\mathbf{x}_{h, v}\in \mathbb{R}^{K\times D}$ where $K = \left\lfloor \frac{H + F + l - 1}{2} \right\rfloor$, $D = \left\lfloor \frac{3J + l - 1}{2} \right\rfloor$ and $l$ is the length of the filter. To incorporate the wavelet manifold into our diffusion model, we concatenate the four resulting subbands, so that $\mathbf{y} = \mathtt{DWT}(\mathbf{x}) \in \mathbb{R}^{K \times 4D}$. This approach benefits motion generation by allowing the model to leverage both high- and low-frequency information, thereby improving the granularity of the predictions.

\subsubsection{Wavelet Manifold Diffusion (WMD)}
\label{sec:wmd}

\myPara{Motion Wavelet Manifold Training.}
Our method is a diffusion-based framework, \ie, DDIM~\cite{song2020denoising}. The diffusion model is trained using the motion wavelet manifolds where both low and high-frequency signals are effectively captured for generation.
We denote $\mathbf{y} = \{\mathbf{y}^{(t)}\}_{t=0}^T$ as a trajectory of noised wavelet manifolds, where $\mathbf{y}^{(0)} = \mathtt{DWT}(\mathbf{x})$ is the wavelet-transformed motion sequence $\mathbf{x} \in \mathbb{R}^{(H+F) \times 3J}$. In the forward diffusion process, each $\mathbf{y}^{(t)}$ is obtained by progressively adding Gaussian noise to $\mathbf{y}^{(t-1)}$ according to a predefined noise scheduler $\mathbf{\alpha} = \{\alpha_t\}_{t=0}^T$ ($\alpha_t \in [0, 1]$), which controls the noise level over the diffusion steps.

We denote $\epsilon_\theta(\mathbf{y}^{(t)}, \star, t)$ as predicted noise with condition $\star$ and $\epsilon_\theta(\mathbf{y}^{(t)} , t)$ for unconditional noise prediction. Specifically, in human motion prediction, we treat the observed motion wavelet $\mathbf{y}^{(0)}=\mathtt{DWT}(\mathbf{x}^{1:H})$ as the input condition. Therefore, the objective function for training MotionWavelet is the following noise prediction objective, 
\begin{equation}
L(\theta) = \mathbb{E}_t \left[\|\epsilon - \epsilon_\theta(\mathbf{y}^{(t)}, \mathbf{y}^{(0)}, t)\|^2\right],
\end{equation}
where $\epsilon$ is the injected noise at each step and $\epsilon_{\theta}(\cdot)$ is the noise prediction network TransLinear~\cite{chen2023humanmac}. During training, the conditioning $\mathbf{y}^{(0)}$ is randomly dropped as $\emptyset$. To sample from the data distribution in the reverse process, we obtain each $\mathbf{y}^{(t-1)}$ from $\mathbf{y}^{(t)}$ as, 
\begin{equation}
\small
\mathbf{y}^{(t-1)} = \dfrac{1}{\sqrt{\alpha_t}} \left( \mathbf{y}^{(t)} - \dfrac{1 - \alpha_t}{\sqrt{1 - \bar{\alpha}_t}} \epsilon_\theta(\mathbf{y}^{(t)}, t) \right)
\label{eq:sample_step}
\end{equation}
Following~\cite{chen2023humanmac}, we adopt TransLinear for noise prediction.

\myPara{Wavelet Manifold Sampling.}
During sampling, classifier-free guidance (CFG) is applied for motion prediction. The prediction guided by $\mathbf{y}^{(0)}$ can be computed as, 
\begin{equation}
\hat{\epsilon}_\theta(\mathbf{y}^{(t)}, \mathbf{y}^{(0)}) = (1 + w) \epsilon_\theta(\mathbf{y}^{(t)}, \mathbf{y}^{(0)}) - w \,\epsilon_\theta(\mathbf{y}^{(t)}),
\end{equation}
where $w$ is the guidance scale. Unlike commonly used large guidance scales, we observed that smaller values of $w < 1$ often lead to more accurate predictions while maintaining diversity. This property stems from the explicit modeling of the wavelet manifold to high-frequency noise, which is sufficient to represent fine-grained motion semantics. Larger $w$ values can amplify larger noise in high-frequency bands. Setting $w$ small allows the model to capture fine-grained motion details without introducing extra noise.

\subsection{Guiding Wavelet Denoising Process}

\begin{figure}
    \centering
\includegraphics[width=\linewidth]{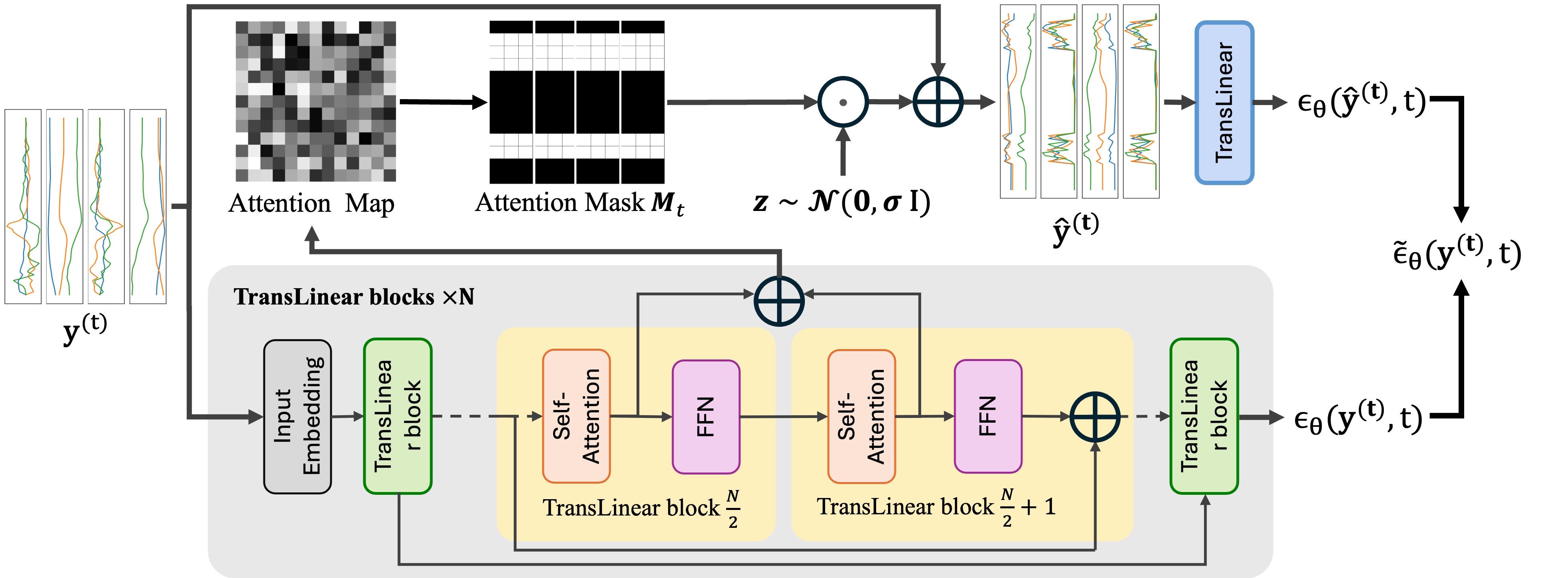}
    \caption{Temporal Attention-Based Guidance~(TABG). The sequence $\y^{(t)}$ is first input to the TransLinear network to obtain an unguided noise prediction. The attention maps from the middle two TransLinear blocks are then summed and averaged to create an aggregated attention mask. This mask is then applied to obtain a noise sequence $\hat{\y}^{(t)}$, which the TransLinear then processes to obtain the TABG-guided noise prediction. The final output is a linear combination of two noise predictions.}
    \label{fig:TABG}
\end{figure}

\begin{algorithm}[!t]
\footnotesize
\caption{WDM Sampling}
\textbf{Functions:}
$\epsilon_\theta$: noise predictor that outputs the predicted noise $\varepsilon_t$, and attention maps $A_t$ given $\mathbf{y}^{(t)}$ at timestep $t$.
\BlankLine
$\mathtt{WMSG}$: Wavelet Manifold Shaping Guidance.

\textbf{Variables:}
noise scale $\sigma$, TABG scaling factor $s$, classifier-free guidance scaling factor $w$, attention threshold $\varphi$,  Conditioning padded history $\mathbf{y}^{(0)}_{\text{pad}}$.
\BlankLine
$\mathbf{y}^{(T)} \sim \mathcal{N}(\mathbf{0}, \mathbf{I})$ \\
\BlankLine
\For{$t = T$ \textbf{to} $1$}{
    \BlankLine
    \quad $\varepsilon_t$, $A_t \leftarrow \epsilon_\theta(\mathbf{y}^{(t)}, t)$, \quad $\varepsilon_{\text{cond}, t} \leftarrow \epsilon_{\theta}(\mathbf{y}^{(t)}, \mathbf{y}^{(0)}_{\text{pad}}, t)$, \quad $M_t[i, :] = \mathbbm{1}_{\{ A_t[i] > \varphi \}}$, \quad $\mathbf{z} \sim \mathcal{N}(\mathbf{0}, \mathbf{I})$ \\
    \BlankLine
    \quad $\tilde{\mathbf{y}}^{(0)} \leftarrow \dfrac{1}{\sqrt{\bar{\alpha}_t}} \left( \mathbf{y}^{(t)} - \sqrt{1 - \bar{\alpha}_t} \varepsilon_t + \sigma \mathbf{z} \right)$ \\
    \BlankLine
    \quad $\tilde{\mathbf{y}}^{(t)} \leftarrow \sqrt{\bar{\alpha}_t} \tilde{\mathbf{y}}^{(0)} + \sqrt{1 - \bar{\alpha}_t} \mathbf{z}$ \\
    \BlankLine
    \quad $\hat{\mathbf{y}}^{(t)} \leftarrow (1 - M_t) \odot \mathbf{y}^{(t)} + M_t \odot \tilde{\mathbf{y}}^{(t)}$ \\
    \BlankLine
    \quad $\tilde{\varepsilon}_t \leftarrow \varepsilon_t + s \left( \epsilon_\theta(\hat{\mathbf{y}}^{(t)}, t) - \varepsilon_t \right)$ \\
    \BlankLine
    \quad $\hat{\varepsilon}_t \leftarrow \tilde{\varepsilon}_t + w(\varepsilon_{\text{cond}, t} - \tilde{\varepsilon}_t)$ \\
    \BlankLine
    \quad $\mathbf{y}^{(t-1)} \leftarrow \mathtt{WMSG}\left(\dfrac{1}{\sqrt{\alpha_t}} \left( \mathbf{y}^{(t)} - \dfrac{1 - \alpha_t}{\sqrt{1 - \bar{\alpha}_t}} \hat{\varepsilon}_t \right)\right)$ \\
    \BlankLine
    
}
\Return $\mathbf{x} = \mathtt{iDWT}(\mathbf{y}^{(0)})$ 
\label{algo: inference}
\end{algorithm}
\paragraph{Wavelet Manifold Shaping Guidance.}
In the iterative denoising process, traditional approaches~\cite{chen2023humanmac, wei2023human} primarily focus on sequentially applying denoising operations to noisy inputs, aiming to gradually reduce noise across timesteps. However, we observe that such methods may not be fully optimal when dealing with complex wavelet manifolds, as the directional trajectory of denoising steps can diverge from the natural structure of wavelet manifolds. This misalignment can hinder the predictive accuracy and efficiency of the denoising process. To address this, we propose the \textit{Wavelet Manifold Shaping Guidance} (WMSG) technique, which integrates a wavelet transform at the end of each iteration to refine the noisy manifold. Specifically, instead of directly applying the subsequent denoising process to the wavelet manifold $\y^{(t-1)}$ obtained from \cref{eq:sample_step}, we first map the output of each step via Inverse Discrete Wavelet Transform (iDWT), $\mathbf{x}' = \mathtt{iDWT}\left(\mathbf{y}^{(t-1)}\right)$, which gives a motion sequence $\mathbf{x}'$. Next, we convert the $\mathbf{x}'$ to wavelet manifold using  Discrete Wavelet Transform~(DWT) $\mathbf{y}^{(t-1)} = \mathtt{DWT}\left(\mathbf{x}'\right)$. The process unfolds as follows, 
\begin{equation}
    \mathbf{y}^{(t-1)} = \mathtt{DWT}\left(\mathtt{iDWT}\left(\mathbf{y}^{(t-1)}\right)\right).
\end{equation}
By shaping the latent space in alignment with the wavelet manifolds, this approach facilitates a more structured progression of denoising steps, enhancing both the quality and fidelity of the generated results. This guidance also effectively stabilizes the noise in the wavelet latent space and smoother transitions across timesteps. We validate the effectiveness of the design in Sec.~\ref{sec:wmsg_guidance}.

\paragraph{Temporal Attention-Based Guidance.}
To further improve the predictive accuracy during the denoising process, we introduce Temporal Attention-Based Guidance~(TABG) during the sampling. An overview of TABG is given in Fig.~\ref{fig:TABG}. Inspired by \cite{hong2023improving}, the attention maps from the middle two layers are taken and averaged across the maps followed by summing along the first dimension to obtain the relative importance of each time step $A_t \in \R^{K}$. An attention mask $M_t \in \R^{K\times D}$ is then generated so that $M_t\left[i - \frac{m-1}{2}: i + \frac{m-1}{2}, \, : \right] = 1$ if $A_t[i] > \varphi$, and $0$ elsewhere. Here $\varphi$ is a masking threshold. Unlike~\cite{hong2023improving}, our masking strategy involves masking neighboring $m$ frames instead of a single frame. This approach enhances the modeling of temporal information crucial for capturing body dynamics. To apply the attention mask, we first obtain a noised version of the intermediate reconstruction of unconditional prediction, 
\begin{equation}
    \tilde{\y}^{(0)} = \hat{\y}^{(0)} + \sigma\mathbf z,
    \label{eq:noised}
\end{equation}
where $\mathbf{z} \sim \mathcal{N}(\textbf{0}, \textbf{I})$, at a predefined noise scale $\sigma$. We then obtain the noisy intermediate timestep $\tilde{\y}^{(t)}$ via posterior sampling on $\tilde{\y}^{(0)}$. Based on the attention mask, we noise the masked area of $\y^{(t)}$ as follows,
\begin{align}
\hat{\y}^{(t)} &= (1 - M_t) \odot \y^{(t)} + M_t \odot \tilde{\y}^{(t)}, \\
\tilde{\epsilon}(\y^{(t)}) &= \epsilon_\theta(\y^{(t)}) + s\left(\epsilon_\theta(\hat\y^{(t)}) - \epsilon_\theta(\y^{(t)})\right), 
\label{eq:TABG_scale}
\end{align}
where $s$ is the TABG scale.  In contrast to~\cite{hong2023improving} where they directly add a new term to the guided prediction of CFG, we propose to combine TABG and CFG by adapting $\tilde{\epsilon}(\y^{(t)})$ as the updated unconditional prediction, and then applying CFG to this new unconditional prediction and the original conditional prediction, which more effectively enhance the sampling performance. The whole sampling process adopting the proposed guidance for motion prediction is detailed in~\cref{algo: inference}.

\section{Experiments}

\begin{table}[!t]
\centering
\caption{Quantitative comparison between our approach and state-of-the-art methods on the HumanEva-I and Human3.6M datasets. Our method consistently demonstrates superior accuracy while maintaining commendable diversity metrics. Bold values indicate the best performance, while underlined values indicate the second best.}
\label{tab:main_res}
\resizebox{\textwidth}{!}{%
\begin{tabular}{lccccccccccc}
\toprule
\multirow{2}{*}{Method} & \multicolumn{5}{c}{HumanEva-I} & \multicolumn{5}{c}{Human3.6M} \\
\cmidrule(lr){2-6} \cmidrule(lr){7-11}
 & APD↑ & ADE↓ & FDE↓ & MMADE↓ & MMFDE↓ & APD↑ & ADE↓ & FDE↓ & MMADE↓ & MMFDE↓ \\
\midrule

ERD~\cite{fragkiadaki2015recurrent} & 0     & 0.382 & 0.461 & 0.521 & 0.595 & 0     & 0.722 & 0.969 & 0.776 & 0.995 \\
acLSTM~\cite{li2017auto}        & 0     & 0.429 & 0.541 & 0.530 & 0.608 & 0     & 0.789 & 1.126 & 0.849 & 1.139 \\
DeLiGAN~\cite{gurumurthy2017deligan}       & 2.177 & 0.306 & 0.322 & 0.385 & 0.371 & 6.509 & 0.483 & 0.534 & 0.520 & 0.545 \\
DSF~\cite{yuan2019diverse}   &4.538 &0.273 &0.290 &0.364 &0.340 & 9.330 & 0.493& 0.592 &0.550& 0.599 \\
BoM~\cite{bhattacharyya2018accurate}           & 2.846 & 0.271 & 0.279 & 0.373 & 0.351 & 6.265 & 0.448 & 0.533 & 0.514 & 0.544 \\
DLow~\cite{yuan2020dlow}         & 4.855 & 0.251 & 0.268 & 0.362 & 0.339 & 11.741 & 0.425 & 0.518 & 0.495 & 0.531 \\
GSPS~\cite{mao2021generating}          & 5.825 & 0.233 & 0.244 & 0.343 & 0.331 & 14.757 & 0.389 & 0.496 & 0.476 & 0.525 \\
MOJO~\cite{zhang2021we}          & 4.181 & 0.234 & 0.260 & 0.344 & 0.339 & 12.579 & 0.412 & 0.514 & 0.497 & 0.538 \\
DivSamp~\cite{dang2022diverse}       & \underline{6.109} & 0.220 & 0.234 & 0.342 & \underline{0.316} & 15.310 & 0.370 & 0.485 & 0.475 & 0.516 \\
STARS~\cite{xu2022diverse} & 6.031 & \underline{0.217} & 0.241 & \underline{0.328} & 0.321 & \textbf{15.884} & \underline{0.358} & \underline{0.445} & \textbf{0.442} & \underline{0.471} \\

MotionDiff~\cite{wei2023human}    & 5.931 & 0.232 & 0.236 & 0.352 & 0.320 & \underline{15.353} & 0.411 & 0.509 & 0.508 & 0.536 \\
Belfusion~\cite{barquero2023belfusion}     & -     & -     & -     & -     & -     & 7.602  & 0.372 & 0.474 & 0.473 & 0.507 \\
HumanMAC~\cite{chen2023humanmac}      & \textbf{6.554} & \textbf{0.209} & \underline{0.223} & 0.342 & 0.320 & 6.301  & 0.369 & 0.480 & 0.509 & 0.545 \\
CoMotion~\cite{sun2023towards}      & -     & -     & -     & -     & -     & 7.632  & \textbf{0.350} & 0.458 & 0.494 & 0.506 \\
\midrule \specialrule{0em}{0.0pt}{0.0pt} 
\rowcolor{lightgray} \name       & 4.171 &0.235 & \textbf{0.213}  & \textbf{0.304}  & \textbf{0.280}  & 6.506 & 0.376 & \textbf{0.408} & \underline{0.466} & \textbf{0.443}\\
\specialrule{0em}{0.0pt}{0.0pt} \bottomrule
\end{tabular}%
}

\end{table}
\begin{figure}
    \centering
    \includegraphics[width=\linewidth]{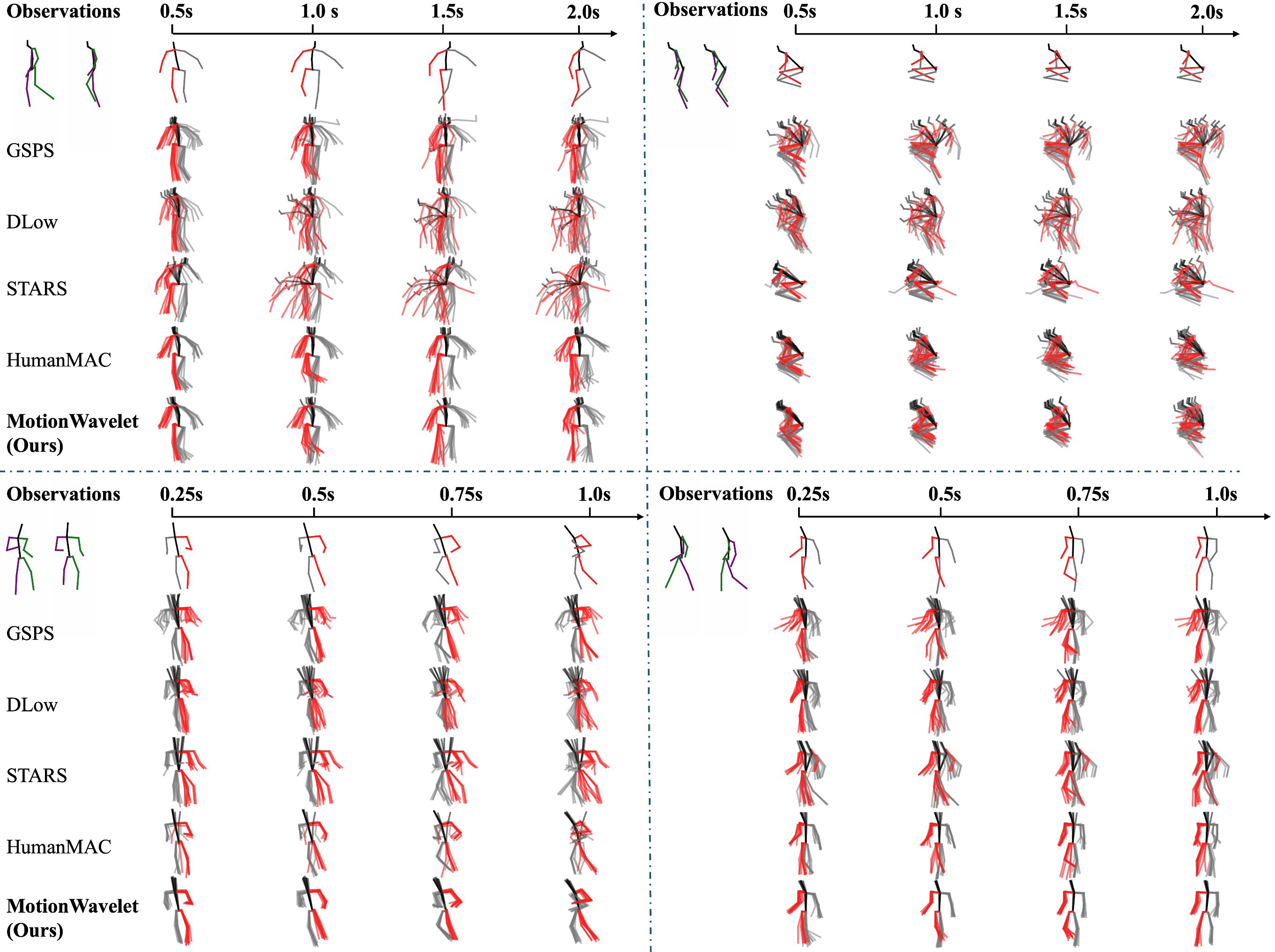}
    \caption{
    Qualitative comparisons. The upper part shows predictions for Human3.6M\cite{ionescu2013human3}, and the bottom part for HumanEva-I\cite{sigal2010humaneva}. The first row in each part represents ground truth motion. The closer to the ground truth motion indicates better prediction.}
    \label{fig:mainresults}
\end{figure}

\subsection{More Qualitative Results}
\label{supp:more_qualitative_res}
\begin{figure}
\vspace{-1mm}
    \centering
    \includegraphics[width=\linewidth]{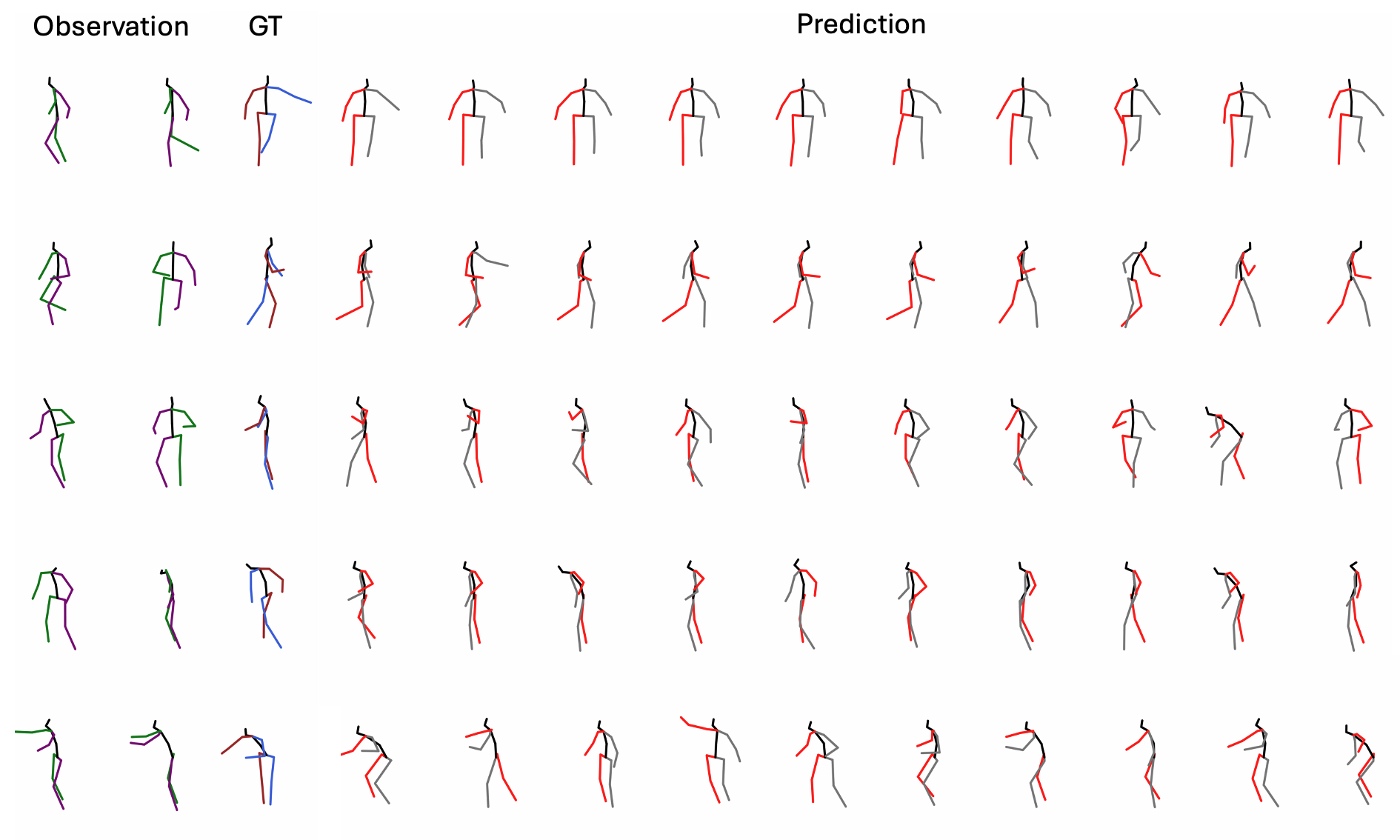}
\vspace{-2mm}
    \caption{
More qualitative results of \name, where the green-purple skeletons represent the observed motions, the blue-purple skeletons represent the GT motions, and the red-black skeletons represent the predicted motions. We visualize 10 predicted samples without overlay.}
\label{fig:more_vis_no_overlay}
    \vspace{-.5em}
\end{figure}

\begin{figure}
\vspace{-10mm}
    \centering
    \includegraphics[width=\linewidth]{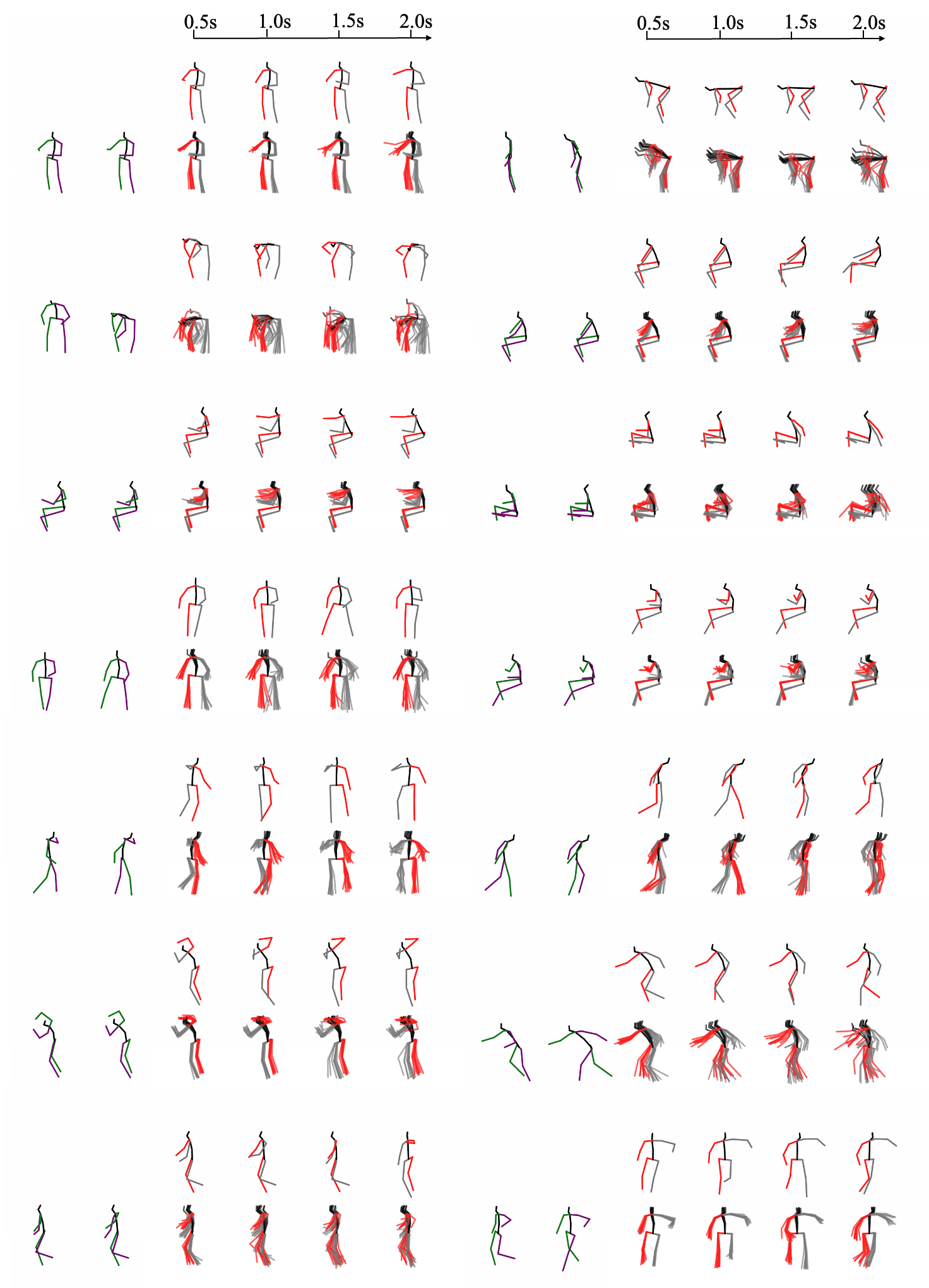}
\vspace{-1mm}
    \caption{
More qualitative results of \name, where the green-purple skeletons represent the observed motions, and the red-black skeletons represent the predicted motions. We visualize 10 predicted samples. Our method produces high-fidelity and diverse motion prediction results.\label{fig:more_vis_overlay}
}
    \vspace{-1.5em}
\end{figure}

In the following, we present additional qualitative results of \name. Fig.~\ref{fig:more_vis_overlay} provides a comprehensive visualization of the ground truth (GT) motions alongside our predictions at each time step, demonstrating the high quality and accuracy of our model. Furthermore, Fig.~\ref{fig:more_vis_no_overlay} showcases 10 predicted motion samples and the end pose of the GT motions based on the observed motion frames. These results highlight the ability of \name to generate diverse motion predictions that align well with the GT while exhibiting notable motion diversity.

\subsection{Datasets}
We present our results on two widely recognized datasets.\\
\noindent \textbf{HumanEva-I dataset}~\cite{sigal2010humaneva} features recordings from 3 subjects at a frequency of 60Hz. Each participant demonstrates 5 actions, represented by a skeleton with 15 joints. Here, we aim to predict 60 future frames (1 second) based on 15 observed frames (0.25 seconds).\\
\noindent \textbf{Human3.6M dataset}~\cite{ionescu2013human3} comprises 3.6 million video frames captured from 11 individuals, with 7 of them providing ground truth data. Each participant executes 15 distinct actions, and the motion data is recorded at a frequency of 50 Hz. For training our model, we utilize data from 5 subjects (S1, S5, S6, S7, S8), while the remaining subjects (S9, S11) are reserved for evaluation. In our analysis, we focus on a skeleton structure consisting of 17 joints for each frame, ensuring the removal of global translation effects. Our prediction involves generating 100 future frames (equivalent to 2 seconds) based on 25 observed frames (0.5 seconds).

\subsection{Metrics}
Following~\cite{chen2023humanmac, xu2025learning, mao2019learning, sun2023towards}, we utilize five metrics to assess our model's performance: \textit{Average Pairwise Distance (APD)} measures the L2 distance between all motion examples, serving as an indicator of result diversity. \textit{Average Displacement Error (ADE)} is defined as the minimum average L2 distance between the ground truth and predicted motion, reflecting the accuracy of the entire predicted sequence. \textit{Final Displacement Error (FDE)} represents the L2 distance between the prediction results and the ground truth in the final prediction frame. \textit{Multi-Modal-ADE (MMADE)} extends ADE to a multi-modal context, where ground truth future motions are categorized based on similar observations. \textit{Multi-Modal-FDE (MMFDE)} follows suit as the multi-modal counterpart to FDE.

\subsection{Implementation Details}
The \name is trained using 1,000 noising steps and the DDIM sampler is set to 100 steps on both datasets. For Human3.6M, the noise prediction network consists of 12 TransLinear blocks, for HumanEva-I it consists of 6 TransLinear blocks. Each TransLinear block has a latent dimension of 768. Both models were trained for 500 epochs with a batch size of 64, and EMA decay was applied throughout the training. We used the AdamW optimizer with an initial learning rate of $1\times 10^{-4}$. All experiments were conducted on the Nvidia RTX A6000 GPU. More details can be found in our supplementary material.

\begin{figure}[!t]
    \centering
    \captionsetup[subfigure]{aboveskip=-2pt, belowskip=0pt, font=small}
    \begin{subfigure}[b]{0.48\textwidth}
        \centering
\includegraphics[width=1.0\linewidth]{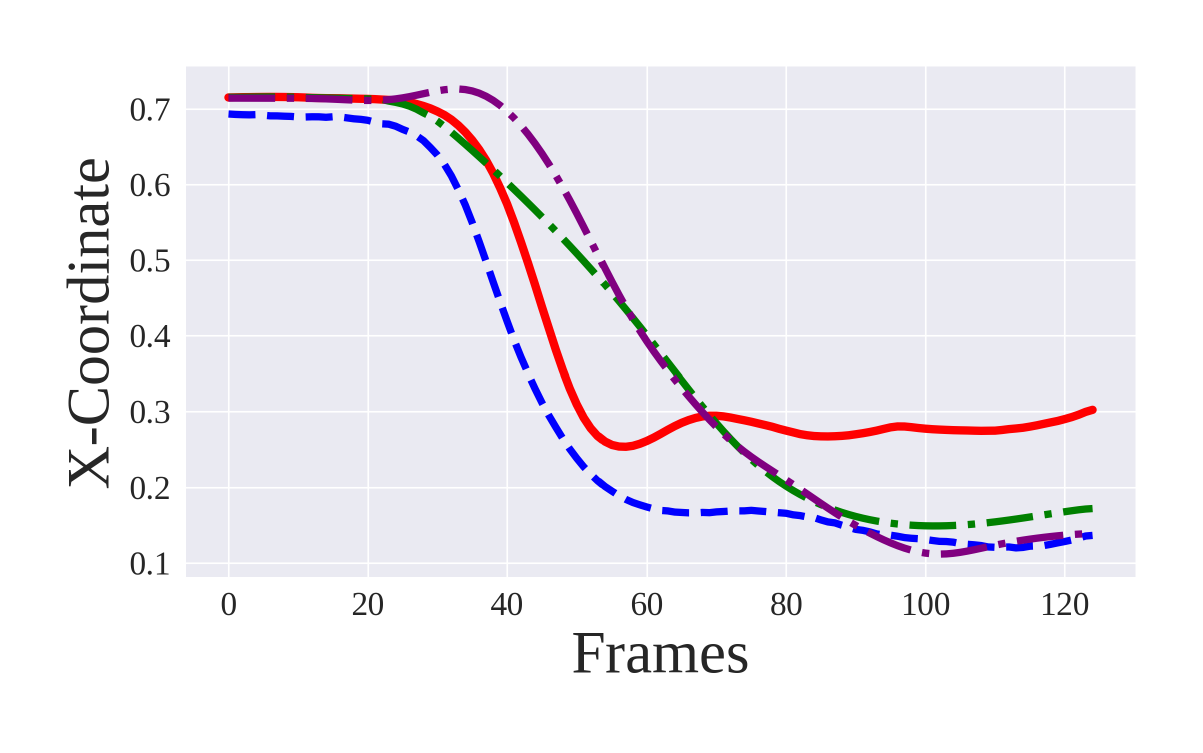}
        \caption{Comparison of left wrist $x$-axis in ``Walking Dog''.}
    \end{subfigure}
    \quad
    \begin{subfigure}[b]{0.48\textwidth}
        \centering
        \includegraphics[width=1.0\linewidth]{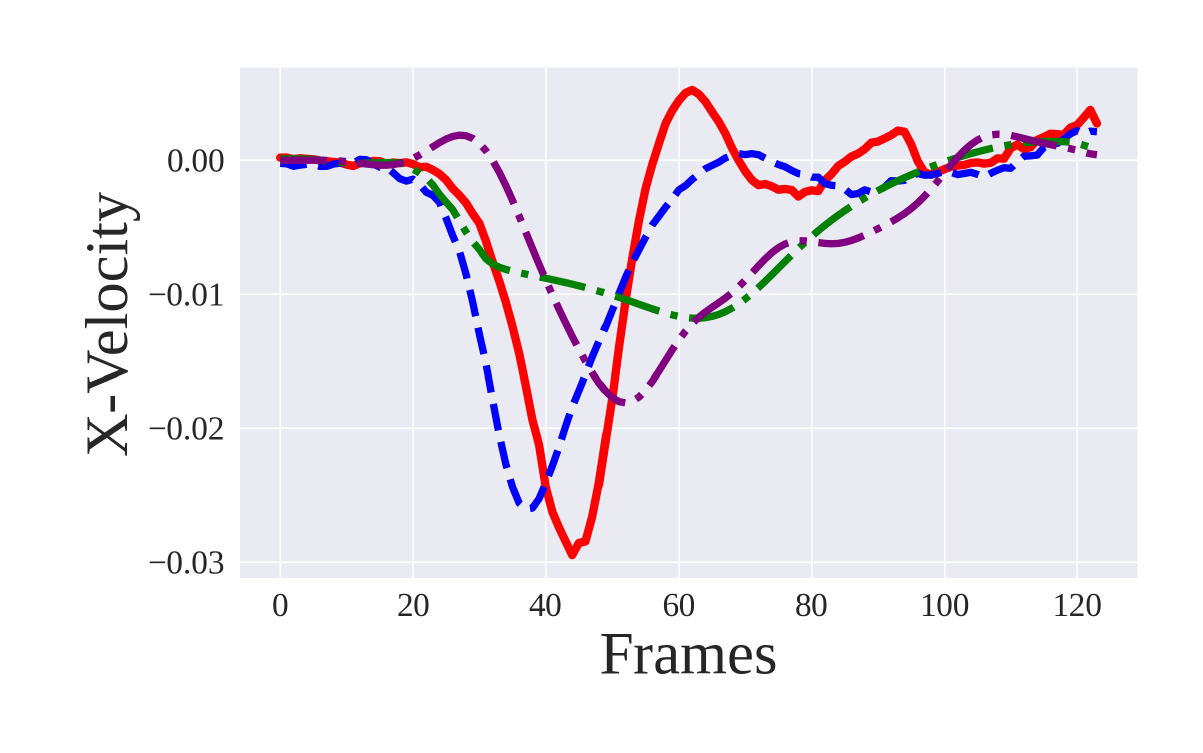}
        \caption{Comparison of left wrist $x$-vel.in ``Walking Dog''.}
    \end{subfigure}
    \begin{subfigure}[b]{0.48\textwidth}
        \centering
\includegraphics[width=1.0\linewidth]{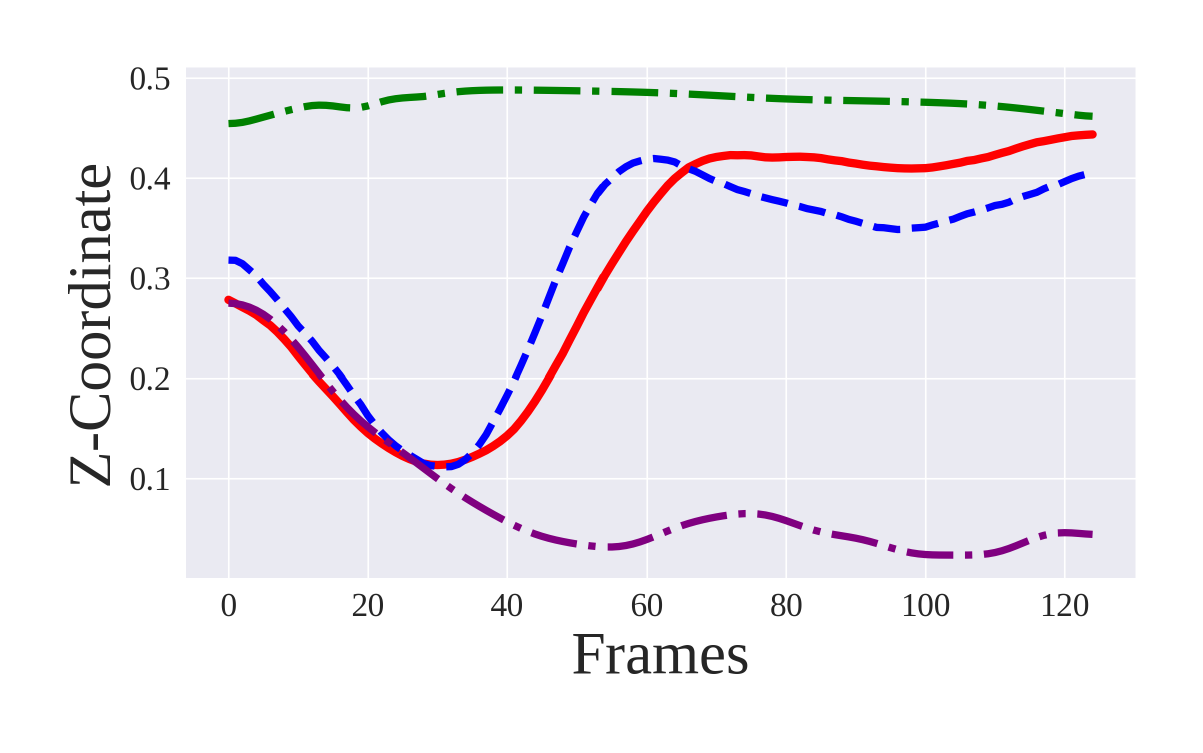}   
        \caption{Comparison of right arm $z$-axis in ``Discussion''.}
    \end{subfigure}
        \quad
    \begin{subfigure}[b]{0.48\textwidth}
        \centering
\includegraphics[width=1.0\linewidth]{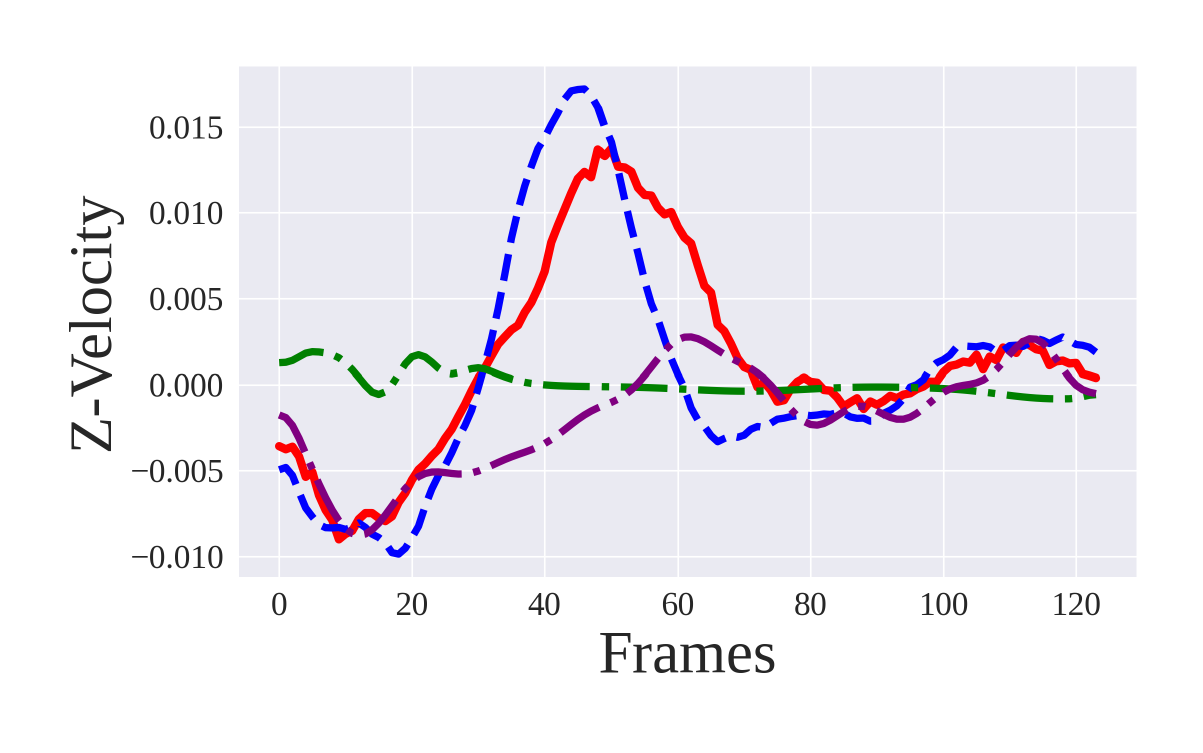}
        \caption{Comparison of right arm $z$-vel. in ``Discussion''.}
    \end{subfigure}
    \caption{Visualization of GT and predicted motion curves of the left wrist for ``Walking Dog'' and the right arm for ``Discussion'' in Human3.6M (vel. denotes velocity). The red curve and the blue line represent GT motion and \name prediction. The purple line represents HumanMAC, and the green line represents DLow. \name achieves better alignment with the ground truth motion.}
    \label{fig:sudden}
\end{figure}

\subsection{Main Results}
\subsubsection{Motion Prediction}
We perform extensive comparisons with state-of-the-art methods of human motion prediction. In Tab.~\ref{tab:main_res}, \name achieves the \textit{overall} best performance. Specifically, our approach achieves the highest performance on the HumanEva-I~\cite{sigal2010humaneva} dataset in terms of FDE, MMADE, and MMFDE, while also demonstrating competitive diversity in motion predictions, as indicated by APD. On the Human3.6M~\cite{ionescu2013human3} dataset, our method continues to exhibit strong performance across multiple metrics, achieving notable accuracy and fidelity alongside commendable diversity. Notably, an increase in APD does not always indicate higher prediction quality, as this metric can rise even with low-quality, disorganized predicted motions. In Fig.~\ref{fig:mainresults}, we qualitatively compare our method with GSPS~\cite{mao2021generating}, DLow~\cite{yuan2020dlow}, STARS~\cite{xu2022diverse}, and HumanMAC~\cite{chen2023humanmac}, where our method consistently shows better visualization results.\\

\paragraph{Motion Analysis.}
In \cref{fig:endposes}, we present various end poses generated by our model in response to the same input observation, illustrating the \textit{diversity} of predicted motion. Additionally, to demonstrate the efficacy of the wavelet manifold in motion prediction, \cref{fig:sudden} examines cases with abrupt transitions—such as sudden stops and starts—providing insight into the ability to capture dynamic features. The result highlights our model's \textit{responsiveness} to diverse motion patterns, demonstrating its robustness in capturing complex and varied movements.

\subsection{Controllable Human Motion Prediction}
\label{supp:controllable} 
During Wavelet Manifold Shaping Guidance (WMSG), as detailed in Sec. 3.3 of the main paper, \name achieves \textit{Controllable Motion Prediction} by blending the noisy ground truth motion with the predicted noisy motion using a predefined temporal or spatial mask, enabling control both at the joint level and during motion transitions. Specifically, we first generate the noisy ground truth motion wavelet manifold, $\y_{\text{gt}}^{(t-1)}$ via the forward diffusion process on the original ground truth motion,
\begin{equation}
    {\y_{\text{gt}}}^{(t-1)} = \sqrt{\bar{\alpha}_{t-1}} {\y}_{\text{gt}}^{(0)} + \sqrt{1 - \bar{\alpha}_t} \mathbf{z}
\end{equation}
where $\mathbf{z} \sim \mathcal{N}(\mathbf{0}, \mathbf{I})$. Then, we use a mask $M\in \R^{(H+F)\times 3J}$ and the predicted noisy motion wavelet manifold $\y^{(t-1)}$ obtained immediately before the final WMSG stage, the controlled motion could be achieved by introducing extra masking within the WMSG process,
\begin{equation}
    \x^{'}_{\text{c}} = (1 - M) \odot \mathtt{iDWT}\left(\y^{(t-1)}\right) + M \odot \mathtt{iDWT}\left(\y_{\text{gt}}^{(t-1)}\right).
    \label{eq:mask_WMSG}
\end{equation}
where $\x^{'}_{\text{c}}$ represents the controlled motion. We then apply DWT to transform the controlled motion into wavelet manifold $\y^{(t-1)} = \mathtt{DWT}( \x^{'}_{\text{c}})$.

In the following, we showcase two applications of our model which are Joint-level Control~ (Sec.~\ref{supp_sub:joint}) and Motion Switch Control~(Sec.~\ref{supp_sub:motoin_switch_control}).

\subsubsection{Joint-level Control}
\label{supp_sub:joint}
\begin{figure}
\vspace{-13mm}
    \centering
    \includegraphics[width=1.01\linewidth]{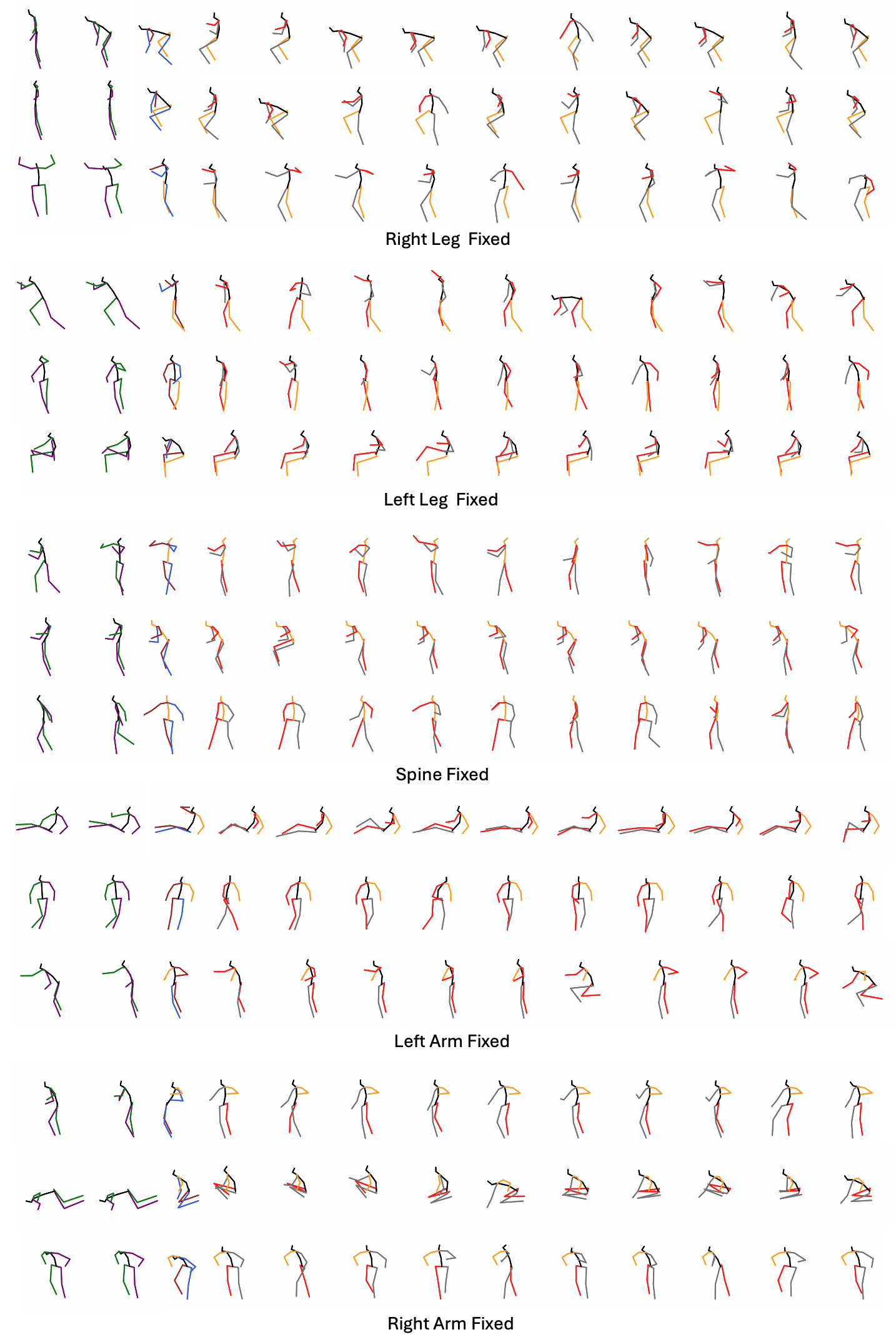}
    \vspace{-3mm}
    \caption{
Visualizations showcasing the joint-level control motion prediction results of \name. The green-purple skeletons represent the observed joint motions, while the red-black skeletons represent 10 end poses of the predicted motions. The controlled joints are highlighted in yellow for clarity. The GT motions are in blue-purple.}
    \label{fig:motion_switch}
    \vspace{-1.5em}
\end{figure}

Given a set of joints coordinates \(C\) containing the indices of the joints to be controlled, the mask \(M\) for Joint-level Control is defined as, 
\[
M[:, j] =
\begin{cases} 
1, & \text{if } j \in C, \\
0, & \text{otherwise.}
\end{cases}.
\]
Substituting this definition into the masking equation \cref{eq:mask_WMSG}, we achieve joint-level control by smoothly blending the predicted and ground truth motions. Next, we showcase that our pipeline could achieve flexible motion prediction where the user can control the specific joints during the motion prediction. The results are shown in \cref{fig:motion_switch}, where we illustrate joint-level controllable motion predictions by specifying right leg, left leg, spine, left arm, and right arm.

\subsubsection{Motion Switch Control}
\label{supp_sub:motoin_switch_control}
\begin{figure}[t]
\vspace{-13mm}
    \centering
    \includegraphics[width=\linewidth]{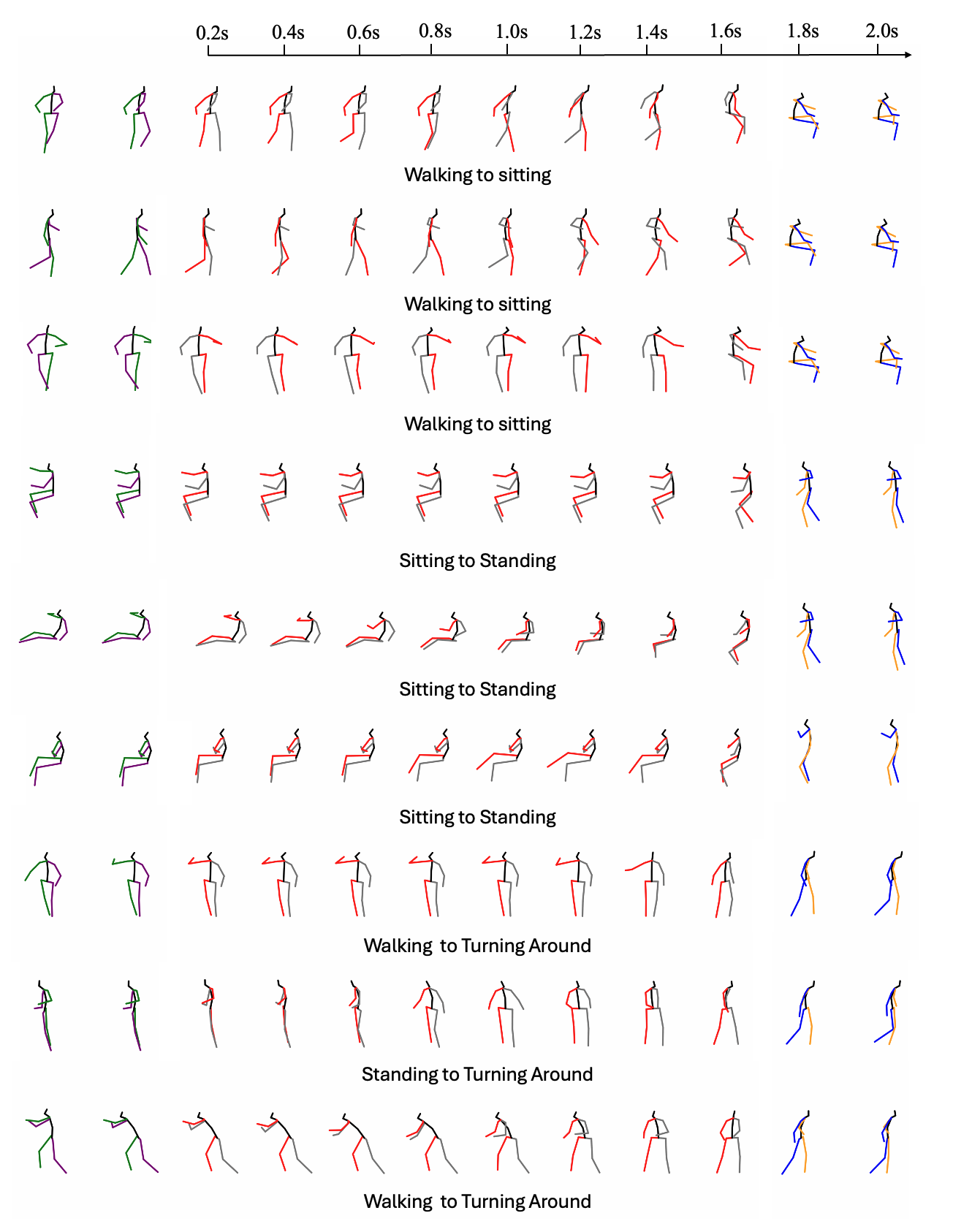}
    \vspace{-0.15em}
    \caption{
Controllable Motion Prediction: Motion Switching. Visualizations showcasing the motion transfer results of \name. The green-purple skeletons represent the observed motions, the red-black skeletons represent the predicted motions, and the blue-yellow skeletons represent the target motions.
}
    \label{fig:motion_trans1}
    \vspace{-1.5em}
\end{figure}

\begin{figure}[t]
\vspace{-13mm}
    \centering
    \includegraphics[width=\linewidth]{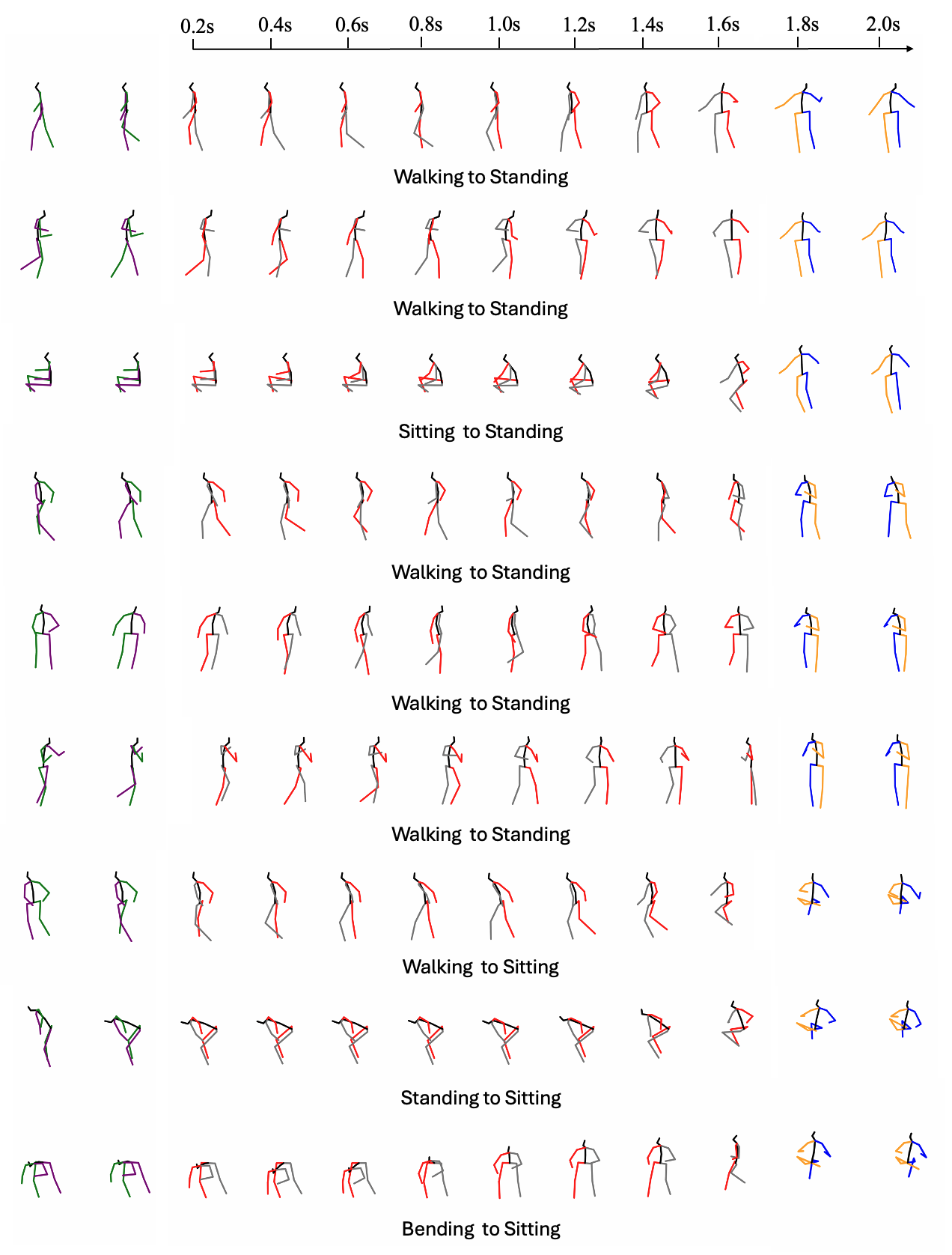}
    \vspace{-0.15em}
    \caption{
Controllable Motion Prediction: Motion Switching. Visualizations showcasing the motion transfer results of \name. The green-purple skeletons represent the observed motions, the red-black skeletons represent the predicted motions, and the blue-yellow skeletons represent the target motions.
}
    \label{fig:motion_trans2}
    \vspace{-1.5em}
\end{figure}

Motion switching can be achieved by specifying a set of frame indices $C$, which represents the timesteps where different motions are to be inserted. The corresponding mask is then defined as,
\[
M[t, :] =
\begin{cases} 
1, & \text{if } t \in C, \\
0, & \text{otherwise.}
\end{cases}
\]
Similarly, by substituting $M$ into \cref{eq:mask_WMSG}, we could dynamically blend motions, and facilitate seamless transitions between different motions at the desired frames.
In Fig.~\ref{fig:motion_trans1} and Fig.~\ref{fig:motion_trans2}, we show more visualization results where our model produces high-quality motion prediction results given the input observation frames and target motion frames. Our model successfully generates smooth and realistic motion transitions between observations and targets, maintaining natural continuity across the sequence. Even in challenging scenarios involving substantial movements or abrupt transitions, such as the transition from Walking to Turning Around, our model still exhibits superior motion coherence and minimal artifacts.

\clearpage

\subsection{Ablation Study}

\subsubsection{Wavelet Space \vs Phase Space}
\label{sec:wavelet_vs_phase}

\begin{figure}[t]
    \centering
    \begin{minipage}{0.42\textwidth}
        \includegraphics[width=\linewidth]{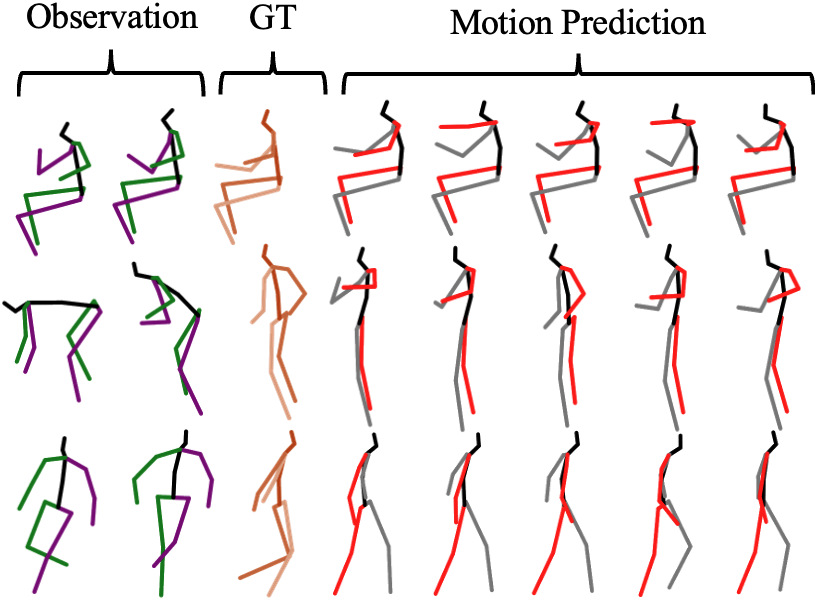}
            \vspace{-1em}
            \caption{Diverse predicted end poses of \name.}
            \label{fig:endposes}
            \end{minipage}
    \hfill
    \begin{minipage}{0.54\textwidth}
            \includegraphics[width=\linewidth]{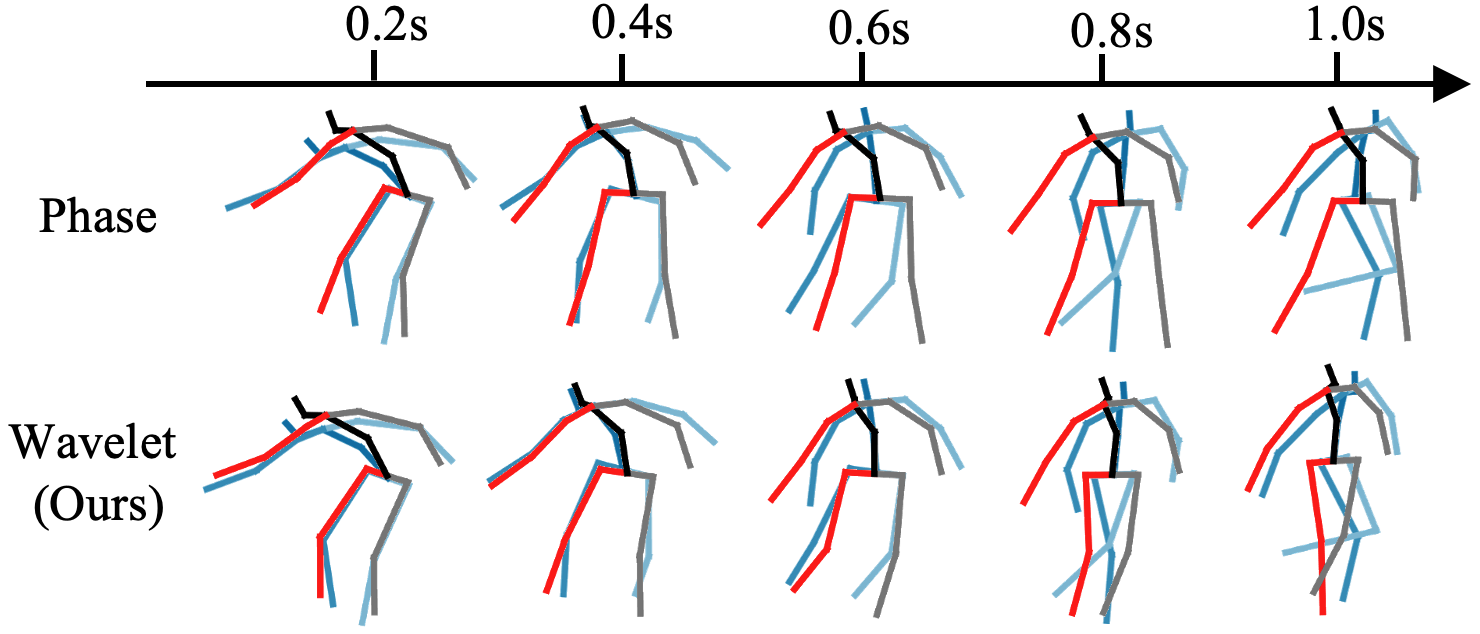}
            \caption{Comparison of prediction results between wavelet and phase for motion representation. The blue skeletons are the ground truth poses and the black and red skeletons are the predicted poses.}
            \label{fig:phase_vs_wavelet}
    \end{minipage}
\end{figure}

\begin{wraptable}{r}{0.4\textwidth}
    \vspace{-1em}
    \centering
    \setlength{\tabcolsep}{3pt}
    \small
    \caption{Phase \vs Wavelet for motion representation learning.}
    \label{tab:phase_vs_wavelet}
    \vspace{-0.3em}
    \resizebox{0.4\textwidth}{!}{
    \begin{tabular}{ccccc}
        \toprule
        &RMSE$\downarrow$& APD$\uparrow$ & ADE$\downarrow$ &FDE$\downarrow$ \\
        \midrule
        Phase & $6.526 \times 10^{-3}$ & 4.925 & 0.619& 0.658 \\ 
        \rowcolor{lightgray}        
        Wavelet & $\mathbf{3.572 \times 10^{-8}}$ & \textbf{6.301}  & \textbf{0.369} & \textbf{0.480}  \\ 
        \bottomrule
    \end{tabular}
    }
    \vspace{-0.5em}
\end{wraptable}
We assess the effectiveness of phase and wavelet representations for motion modeling in the frequency domain. While phase space captures cyclic patterns suitable for periodic motion, it struggles with complex dynamics. In contrast, wavelet space provides a multi-resolution approach, decomposing motion signals across time and frequency to capture localized and non-stationary features more effectively. This foundation better models motion features for diffusion-based motion prediction within the wavelet manifold. As shown in~\cref{tab:phase_vs_wavelet}, the wavelet representation consistently achieves higher performance across metrics. The advantages of the wavelet representation are further illustrated in \cref{fig:phase_vs_wavelet}, where our model generates motions more closely aligned with the ground truth.

\subsubsection{Ablation on Different Wavelet Bases}
\label{sec:wavelet_bases}

\begin{table}
    \centering
    \caption{Ablation study on different wavelet bases.}
    \label{tab:wavelet_bases}
    \resizebox{\textwidth}{!}{
    \setlength{\tabcolsep}{0.1pt}
    \small
    \begin{tabular}{cccccccccccc}
        \toprule
       Bases & Bior2.8 & Rbio2.8 & Bior6.8 & Sym9 &Sym10  &  Coif3 & Db9 & Coif5 & Haar &  Dmey \\
        \midrule
        Position RMSE     $\downarrow$ & $\mathbf{3.572 \times 10^{-8}}$ & $4.228 \times 10^{-8}$ & $4.237 \times 10^{-8}$ & $4.390 \times 10^{-8}$ & $4.507 \times 10^{-8}$  & $4.506 \times 10^{-8}$ & $4.591 \times 10^{-8}$ & $5.578 \times 10^{-8}$ & $7.070 \times 10^{-5}$ & $9.825 \times 10^{-4}$ \\
        Velocity RMSE       $\downarrow$ & $\mathbf{3.816 \times 10^{-8}}$ & $5.318 \times 10^{-8}$ & $4.893 \times 10^{-8}$ & $5.677 \times 10^{-8}$ & $5.477 \times 10^{-8}$  & $5.262 \times 10^{-8}$ & $5.315 \times 10^{-8}$ & $6.241 \times 10^{-8}$ & $6.277 \times 10^{-5}$ & $3.784 \times 10^{-5}$ \\
        Acceleration RMSE   $\downarrow$ & $\mathbf{6.307 \times 10^{-8}}$ & $9.112 \times 10^{-8}$ & $8.153 \times 10^{-8}$ & $9.514 \times 10^{-8}$ & $9.116 \times 10^{-8}$  & $8.665 \times 10^{-8}$ & $8.624 \times 10^{-8}$ & $1.025 \times 10^{-7}$ & $8.877 \times 10^{-5}$ & $4.265 \times 10^{-5}$ \\
        \bottomrule
    \end{tabular}
    }
\end{table}
\textit{Which wavelet bases are best suitable for motion embedding?} Next, we provide a comprehensive study of different wavelet bases for motion wavelet manifold learning (Sec.~\ref{sec:waveletmanifold}). Our extensive experiments in Tab.~\ref{tab:wavelet_bases} reveal that Bior2.8 yields the best performance for motion learning, likely due to its symmetry and better expressiveness after the compaction, which capture detailed motion patterns effectively. Its multi-resolution mode isolates subtle features across scales, enhancing accuracy in complex, non-stationary motion sequences.

\subsubsection{Wavelet Manifold Shaping Guidance}
\label{sec:wmsg_guidance}
We validate Wavelet Manifold Shaping Guidance (WMSG) in Tab.~\ref{tab:guidance}. When equipped with WMSG, our model demonstrates enhanced diversity, indicated by a higher APD value, alongside improved accuracy and fidelity across other metrics. Without WMSG, the na\"ive denoising process inadvertently disrupts the underlying wavelet manifold structure, limiting predictive performance. The results validate the effectiveness of WMSG in guiding the denoising process along the wavelet manifold.

\begin{table}[t]
    \centering
    \setlength{\tabcolsep}{13pt}
    \small
    \caption{Ablation study on Wavelet Manifold Shaping Guidance~(WMSG), Temporal Attention-Based Guidance~(TABG) scale $s$ and noise level $\sigma$.}
    \vspace{0.5em}
    \label{tab:guidance}
    \resizebox{0.97\textwidth}{!}{%
    \begin{tabular}{cccccccc}
        \toprule
         $s$ & $\sigma$ &WMSG& APD$\uparrow$ & ADE$\downarrow$ &FDE$\downarrow$ & MMADE$\downarrow$ & MMFDE$\downarrow$ \\
        \midrule
        0& 0 &\xmark & 6.170 & 0.377 & 0.419 & 0.470 & 0.458 \\
        0& 0 &\cmark & \textbf{6.829} & 0.379 & 0.410 & 0.471 & 0.445\\
        \hdashline %
        0.5 & 0.5 &\cmark & 6.795 & 0.378 & 0.409 & 0.471 & 0.444 \\
        0.5 & 1.5 &\cmark & 6.690 & 0.377 & \textbf{0.408} & 0.469 & \textbf{0.443} \\
        0.5 & 2.5 &\cmark & 6.640 & \textbf{0.376} & \textbf{0.408} & 0.468 & \textbf{0.443} \\ 
                \hdashline %
        1.0 & 0.5 &\cmark & 6.780 & 0.379 & 0.409 & 0.471 & 0.444 \\
        1.0 & 1.5 &\cmark & 6.588 & 0.377 & 0.409 & 0.467 & \textbf{0.443} \\
        \rowcolor{lightgray}
        1.0 & 2.5 &\cmark & 6.506 & \textbf{0.376} & \textbf{0.408} & \textbf{0.466} & \textbf{0.443} \\
                    \hdashline %
        1.5 & 0.5 &\cmark & 6.765 & 0.379 & 0.410 & 0.470 & 0.445 \\
        1.5 & 1.5 &\cmark & 6.493 & 0.377 & 0.409 & \textbf{0.466} & \textbf{0.443} \\
        1.5 & 2.5 &\cmark & 6.407 & \textbf{0.376} & 0.410 & \textbf{0.466} & 0.444 \\
        \bottomrule
    \end{tabular}
    }
\end{table}

\subsubsection{Temporal Attention-Based Guidance}
\label{sec:tabg_guidance}

\begin{figure}[t]
    \centering
    \captionsetup[subfigure]{aboveskip=0pt, belowskip=-2pt, font=small}
    \begin{subfigure}[b]{0.45\linewidth}
        \centering
        \includegraphics[width=1.0\linewidth]{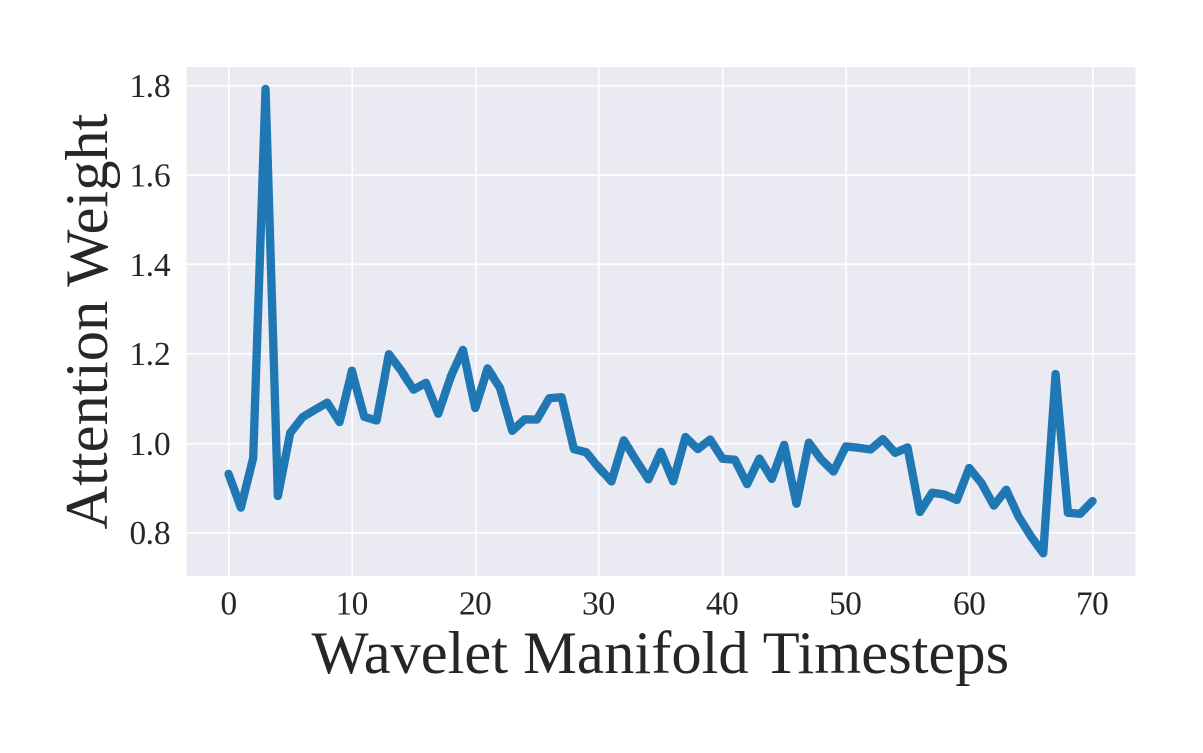}
        \caption{Attention vectors of every frame in the wavelet manifold.}
        \label{fig:attn_sum}
    \end{subfigure}
    \hfill
    \begin{subfigure}[b]{0.45\linewidth}
        \centering\includegraphics[width=1.0\linewidth]{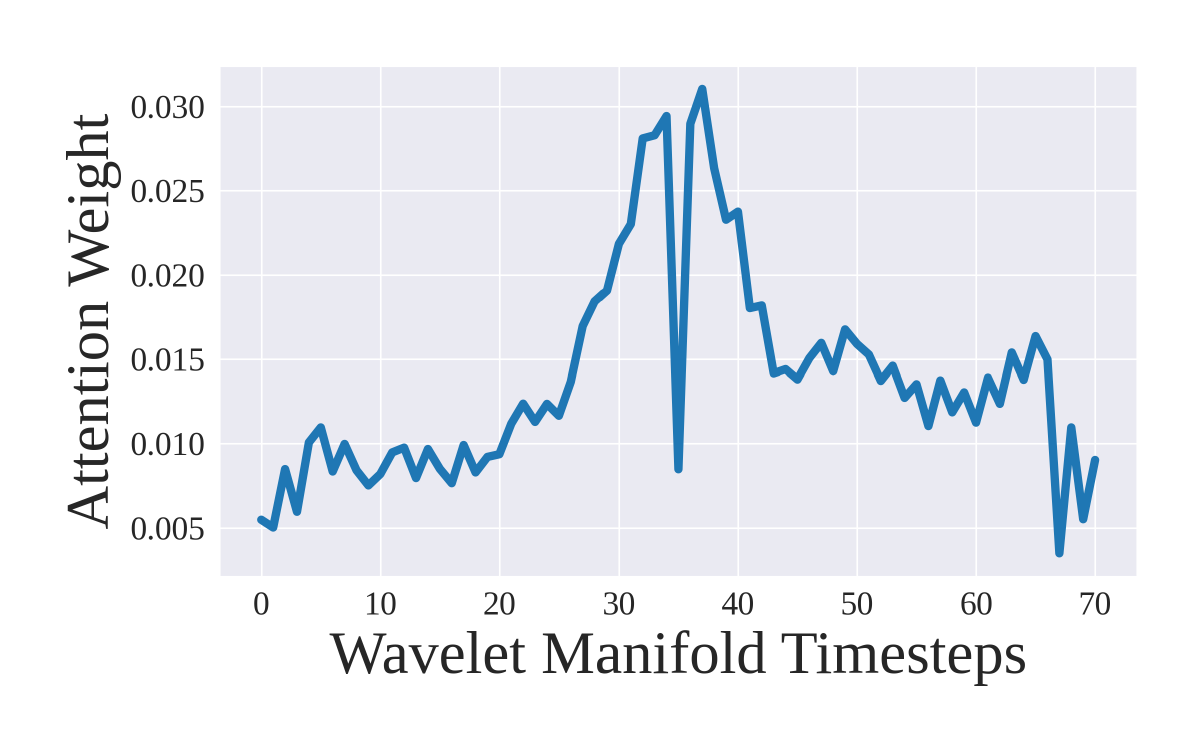}
        \caption{Attention vectors of the 35th frame in the wavelet manifold.}   
        \label{fig:attn_35}
    \end{subfigure}
    \vspace{-0.3em}
    \caption{Visualization of the attention vectors. The horizontal axis represents the wavelet-transformed temporal dimension, while the vertical axis represents the attention weight.}
    \label{fig:attns}
    \vspace{0.5em}
\end{figure}

As shown in Tab.~\ref{tab:guidance}, incorporating Temporal Attention-Based Guidance (TABG) improves prediction accuracy by guiding the denoising process to prioritize key motion features in the frequency domain along the temporal dimension. We evaluate our model with varying noise levels $(\sigma)$ and TABG scales $s$ in Eq.~\ref{eq:noised} and Eq.~\ref{eq:TABG_scale}, respectively. Our extensive experiments indicate that $s=1.0$ and $\sigma =2.5$ yield optimal performance. In \cref{fig:attns}, we visualize the attention map across the wavelet-transformed temporal dimension to analyze the model's temporal dependencies. \cref{fig:attn_sum} displays the overall accumulated attention weight for each frame, demonstrating that the model actively attends to history frames, as these frames receive higher attention scores in the motion sequence. Moreover, \cref{fig:attn_35} shows that the attention vectors for the middle frames focus more on their neighboring frames, suggesting that the model aims to construct more fluent motions.

\subsubsection{Settings of Diffusion model}
In Tab.~\ref{tab:diffusion_model}, we examine the impact of different configurations in the Wavelet Manifold Diffusion (WMD) model by testing various noise and DDIM step settings during training and inference. Results indicate that using a relatively high number of steps for both training and testing enhances model performance. In Tab.~\ref{tab:scheduler}, we evaluate the effect of different schedulers on WMD performance, finding that the Cosine scheduler yields the best results for motion prediction.

\begin{table}[!t]
    \begin{minipage}{0.45\textwidth}   
        \centering
        \setlength{\tabcolsep}{3pt}
        \small
        \caption{Experimental results of the ablation study on different schedulers in the Wavelet Manifold Diffusion model.}
            \vspace{2mm}
            \label{tab:diffusion_model}
                    \resizebox{\textwidth}{!}{%
        \begin{tabular}{ccccc}
            \toprule

            \# Noise steps & \# DDIM steps &  APD$\uparrow$ & ADE$\downarrow$ &FDE$\downarrow$ \\
            \midrule
            100 & 10 & 6.310 & 0.399 & 0.458 \\
            \rowcolor{lightgray}
            1000 & 100 & \textbf{6.506}&  \textbf{0.376}& \textbf{0.408} \\
            \bottomrule
        \end{tabular}
        }
    \end{minipage}
    \hfill
    \begin{minipage}{0.48\textwidth}   

        \centering
        \setlength{\tabcolsep}{3pt}
        \small
        \caption{Ablation study on different Wavelet Manifold Diffusion Model schedulers.}
                    \vspace{3mm}
        \label{tab:scheduler}
            \resizebox{\textwidth}{!}{%
        \begin{tabular}{cccccc}
            \toprule
    
            Scheduler& APD$\uparrow$ & ADE$\downarrow$ &FDE$\downarrow$ & MMADE$\downarrow$ & MMFDE$\downarrow$ \\
            \midrule
            Linear & 5.370 & 0.378 & 0.415 & 0.458 & 0.445\\
            Sigmoid & \textbf{6.552} & 0.380 & 0.423& 0.471& 0.459\\
            \rowcolor{lightgray}
            Cosine & 6.506 & \textbf{0.376} & \textbf{ 0.408} & \textbf{0.466}  & \textbf{0.443}\\
            \bottomrule
        \end{tabular}
        }

    \end{minipage}

\end{table}
\section{Conclusion}
We present a novel method for human motion prediction, termed \name. By constructing a Wavelet Manifold from motion data and applying wavelet diffusion, our method captures motion features in the spatial-frequency domain, allowing effective modeling of both global and local temporal characteristics, as well as the non-stationary dynamics. We introduce the Wavelet Diffusion Model to model discriminative features from the motion wavelet manifold and improve predictive performance through Wavelet Manifold Shaping Guidance and Temporal Attention-Based Guidance mechanisms. Extensive experiments validate the effectiveness of \name, demonstrating remarkably improved prediction accuracy and generalization across different benchmarks.

\section{Limitations and Future Works}
Despite the advantages of \name, our approach has several limitations. The model's performance may be hindered by the availability of high-quality motion data, impacting generalization. Additionally, ensuring the physical plausibility of the predicted motions poses a challenge. Future research could focus on overcoming these limitations by exploring data enhancement techniques, such as learning from videos~\cite{dou2023tore,li2021hybrik,lin2021end,   cai2024smpler, lin2024motion}. Integrating physical~\cite{jiang2023drop,  yuan2023physdiff, peng2022ase, dou2023c, peng2021amp} or biomechanical~\cite{werling2024addbiomechanics, park2023bidirectional, karatsidis2019musculoskeletal, feng2023musclevae} models could improve the realism of generated motions. 

\bibliographystyle{plain}
\bibliography{main}

\newpage

\appendix
\section*{Appendix}

This appendix covers the following sections:
{Implementation Details} (Sec.~\ref{supp:implementation}); and Mathematical Details for Discrete Wavelet Transform~(Sec.~\ref{supp_sec:math}). 
For a more comprehensive overview, please refer to our supplementary video.

\section{Implementation Details}
\label{supp:implementation}
We provide more implementation details. For the Human3.6M dataset, the latent size is 768 and the feed-forward size is 1536. For HumanEva-I, the latent size is 768 and the feed-forward size is 1024. The number of self-attention heads is set to be 8 for both models. The learning rate scheduler for Human3.6M is set to be a multi-step learning rate scheduler with $\gamma=0.2$ and milestones at epoch number 120, 180, 240 and 300. The learning rate scheduler for HumanEva-I is set to be Cosine Annealing With Decay with $\gamma=0.9$ and $T_{max}=20$. Both models are trained on two Nvidia RTX A6000 GPUs with an effective batch size of 64, and the batch size on each device is 32. During inference, for the Human3.6M dataset, TABG is applied in the first 90 denoising steps, and for the HumanEVA-I dataset, TABG and WMSG are not applied. For the Motion Switch Control and Joint-Level Control, the masking is applied in the first 90 denoising steps. The wavelet function is set to be a biorthogonal wavelet with 2 vanishing moments in the decomposition wavelet and 8 vanishing moments in the reconstruction wavelet, and the padding mode is zero at the boundaries. The decomposition level is set to be 1.

\section{Mathematical Details for Discrete Wavelet Transform}
\label{supp_sec:math}
\subsection{1D Inverse Discrete Wavelet Transform}

Given the approximation coefficients $a[k]$ and detail coefficients $d[k]$, the original signal $x[n]$ can be reconstructed using the inverse DWT as, 
\vspace{-0.5em}
\begin{equation}
    x[n] = \sum\nolimits_k a[k] \, \phi_{k}[n] + \sum\nolimits_k d[k] \, \psi_{k}[n],
\vspace{-0.5em}
    \label{eq:1d-idwt}
\end{equation}
where $\phi_k[n]$ and $\psi_k[n]$ are the scaling and wavelet functions defined as $\phi_k[n] = 2^{1/2} \, \phi[2n - k]$ and $\psi_k[n] = 2^{1/2} \, \psi[2n - k]$, respectively.

Practically, the reconstructed signal is obtained by convolution with up-sampling, 
\vspace{-0.5em}
\begin{equation}
    \small
    x[n] = \sum\nolimits_m h'[n - 2m] \cdot a[m] + \sum\nolimits_m g'[n - 2m] \cdot d[m],
\vspace{-0.5em}
    \label{eq:1d-idwt-conv}
\end{equation}
where $h'[n]$ and $g'[n]$ are the reconstruction low-pass and high-pass filters, respectively.

\subsection{2D Inverse Discrete Wavelet Transform}

Given the four subbands $\mathbf{y}_{h,v}[k_1, k_2]$ for $h,v \in \{L,H\}$, the original signal $\mathbf{x}[i,j]$ can be reconstructed using the inverse DWT as, 
\vspace{-0.5em}
\begin{equation}
    \small
    \mathbf{x}[i,j] = \sum_{h,v \in \{L,H\}} \sum_{k_1}\sum_{k_2} \mathbf{y}_{h,v}[k_1,k_2] \, f'_h[i - 2k_1] \, f'_v[j - 2k_2],
\vspace{-0.5em}
    \label{eq:2d-idwt}
\end{equation}
where $f'_L[n] = h'[n]$ and $f'_H[n] = g'[n]$ are the reconstruction filters corresponding to the low-pass and high-pass filters from the 1-D inverse DWT.

Practically, the reconstruction involves convolution with up-sampling along both dimensions, 
\vspace{-0.5em}
\begin{equation}
    \mathbf{x}[i,j] = \sum_{h,v \in \{L,H\}} \left( \left( \mathbf{y}_{h,v}^{\uparrow_2} \ast f'_h \right) \ast f'_v \right)[i,j],
\vspace{-0.5em}
    \label{eq:2d-idwt-conv}
\end{equation}
where $\mathbf{y}_{h,v}^{\uparrow_2}$ denotes up-sampling of $\mathbf{y}_{h,v}$ by a factor of 2 along both dimensions, and $\ast$ denotes the convolution operation.

\end{document}